\documentclass{article}
\PassOptionsToPackage{numbers,sort,compress}{natbib}
\usepackage[preprint]{neurips_2026}
\usepackage{wrapfig}
\usepackage{neurips_2026}
\usepackage[utf8]{inputenc}
\usepackage[T1]{fontenc}    
\usepackage{hyperref}      
\usepackage{url}           
\usepackage{booktabs} 

\usepackage{microtype}
\usepackage{amsfonts}       
\usepackage{nicefrac}       
\usepackage{microtype}      
\usepackage{xcolor} 
\usepackage{graphicx}
\usepackage{tikz}
\usepackage{amsmath}
\usepackage{amsthm}
\usepackage{amssymb}
\usepackage{algorithm}
\usepackage{algpseudocode}
\newtheorem{theorem}{Theorem}
\newtheorem{remark}{Remark}

\usetikzlibrary{shapes, arrows.meta}
\usepackage{makecell}
\usepackage{booktabs} 
\usepackage{pifont}
\usepackage{todonotes}
\usepackage{graphicx}
\usepackage{subcaption}
\usepackage{bbm}

\usepackage{xparse}

\RenewDocumentCommand{\vspace}{s m}{}

\newcommand{\cmark}{\ding{51}} 
\newcommand{\xmark}{\ding{55}}

  \usepackage{tcolorbox}

\newtcolorbox{mathbox}[1][]{colback=gray!20,  #1}
\tcbset{width=(\linewidth-4mm),
before=,after=\hfill,colframe=black!75!black,colback=black!5!gray}

\newcommand{\Nnodes}{n}
\newcommand{\ndim}{d}
\newcommand{\lrtrandom}{B_t}

\newcommand{\honsetset}{\mathcal{K}}
\newcommand{\adversaryset}{\mathcal{Q}}
\newcommand{\acceptancet}{\mathcal{A}_t}
\newcommand{\allnodesset}{\mathcal{V}}
\newcommand{\longalgoname}{Variance-matched Iterative Step-size Tuning for Adaptive optimization}
\newcommand{\algoname}{\textsf{VISTA}}
\newcommand{\pat}{p_{\eta_t}}
\newcommand{\descentstep}{D_t}
\newcommand{\mse}{\mathsf{MSE}}
\newcommand{\pa}{\mathsf{PA}}
\newcommand{\mset}{\sigma_{\eta_t}^2}
\newcommand{\targetvariance}{\sigma^2_{\text{target}}}
\newcommand{\sigmin}{\sigma_{\text{min}}^2}
\newcommand{\sigmax}{\sigma_{\text{max}}^2}
\newcommand{\pmin}{p_{\text{min}}}

\newcommand{\indicatorsaturate}{\mathbb{I}^{\text{sat}}}
\newcommand{\indicatoraceptance}{\mathbb{I}^{\text{acc}}}
\newcommand{\numberofacceptance}{u}

\title{\algoname: Decentralized Machine Learning in Adversary Dominated Environments}

\author{Anonymous Author(s)}

\author{%
  Hanzaleh Akbari Nodehi \\
  University of Minnesota \\
  \texttt{akbar066@umn.edu} \\
  \And
  Parsa Moradi \\
  University of Minnesota \\
  \texttt{moradi@umn.edu} \\
  \AND
  Soheil Mohajer \\
  University of Minnesota \\
  \texttt{soheil@umn.edu} \\
  \And
  Mohammad Ali Maddah-Ali \\
  University of Minnesota \\
  \texttt{maddah@umn.edu} \\
}

\begin{document}

\maketitle
\vspace{-15pt}
\begin{abstract}
 
 Decentralized machine learning often relies on outsourcing computational tasks, such as gradient evaluations, to untrusted worker nodes. Existing robust aggregation methods, including median-based rules, can mitigate malicious behavior under honest-majority assumptions, where most workers follow the prescribed protocol. However, these methods can fail in permissionless environments in which adversaries may control a majority of the workers.

To address this challenge, we consider an incentive-oriented framework in which workers' reports are \emph{accepted} and \emph{rewarded} only when they are mutually consistent within a prescribed \emph{threshold}. Upon acceptance, the submitted reports are used to estimate the desired computational output. This mechanism transforms the adversary from a pure saboteur into a rational agent that must balance the benefit of increasing the estimation error against the risk of rejection and loss of reward.

We study iterative optimization under this adversary-dominated model, focusing on the problem of minimizing a loss function to train a machine learning model. Unlike one-shot computation, iterative optimization requires long-horizon decision making: the system must tune the acceptance rule to achieve fast and reliable convergence over time. This creates a fundamental tradeoff. Permissive acceptance rules can accelerate early progress but allow substantial adversarial corruption, whereas strict rules improve estimation accuracy at the cost of frequent rejections and slower optimization. We propose \algoname, an adaptive algorithm that dynamically tunes the acceptance threshold based on the observed history. We demonstrate the effectiveness of \algoname\ through numerical evaluations and complement these empirical results with a rigorous convergence analysis. These results show that, with suitable incentive-aware adaptation, adversary-dominated decentralized learning can retain the asymptotic convergence behavior of standard SGD without relying on an honest majority.  To the best of our knowledge, this work is the first to address decentralized learning in an adversary-dominated regimes.

\end{abstract}
\vspace{-18pt}



\section{Introduction}
\vspace{-4pt}

\begin{figure*}[ht]
    \centering
    \begin{subfigure}[t]{0.44\textwidth}
        \centering
        \vspace{-25mm}
\includegraphics[width=\linewidth]{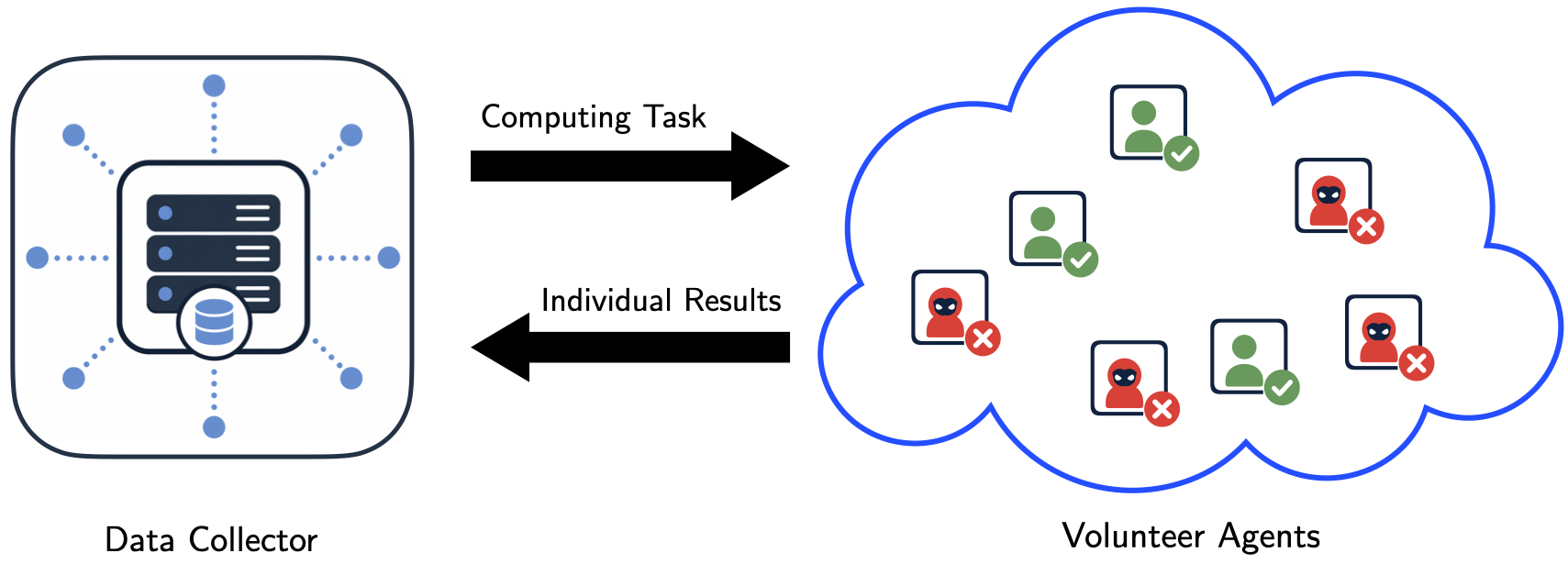}
        \caption{}
        \label{fig:outsourcing_general}
    \end{subfigure}
    \hfill
    \begin{subfigure}[t]{0.55\textwidth}
        \centering
        \resizebox{\linewidth}{!}{\centering
\resizebox{0.95\columnwidth}{!}{%
    \begin{tikzpicture}[>=Stealth, thick]

        \tikzset{
            block/.style={draw, rectangle, minimum height=10mm, minimum width=22mm, align=center, fill=white},
            diamondblock/.style={draw, diamond, aspect=1.6, minimum width=35mm, minimum height=20mm, align=center, fill=white, inner sep=0pt},
            rejectblock/.style={draw, rounded rectangle, minimum height=10mm, minimum width=22mm, align=center, fill=white}
        }

        \def\boxL{0}
        \def\boxR{9.3}
        \def\boxT{2.2}
        \def\boxB{-2.2}
        
        \def\nodeX{-4.3}   
        \def\nodeY{1.4}
        
        \def\diaX{3.0}
        
        \def\estX{7.5}
        \def\rejY{-3}

        \draw[thick, fill=gray!5] (\boxL, \boxT) rectangle (\boxR, \boxB);
        \node[anchor=south east, font=\large] at (\boxR-0.1, \boxB+0.2) {Data Collector};

        \node[block] (node1) at (\nodeX, \nodeY) {Node 1};
        \node[block] (node2) at (\nodeX, -\nodeY) {Node 2};
        
        \node[diamondblock] (decision) at (\diaX, 0) {${||\mathbf{Y}_a - \mathbf{Y}_h||}_2 \leq \eta\Delta$};
        
        \node[block] (estimation) at (\estX, 0) {Estimation};
        
        \node[rejectblock] (reject) at (\diaX, \rejY) {Reject};
        
        \node[font=\Large] (output) at (\boxR + 1.2, 0) {$\hat{L}(\mathbf{U})$};

        
        \draw[->] (node1.east) --
            node[pos=0.52, above=3pt, fill=white, inner sep=1pt]
            {$\mathbf{Y}_h = L(\mathbf{W}) + \mathbf{N}_h$}
            (\boxL, \nodeY);
        \draw[->] (\boxL, \nodeY) -- (decision.155);
        
        \draw[->] (node2.east) --
            node[pos=0.52, above=3pt, fill=white, inner sep=1pt]
            {$\mathbf{Y}_a = L(\mathbf{W}) + \mathbf{N}_a$}
            (\boxL, -\nodeY);
        \draw[->] (\boxL, -\nodeY) -- (decision.205);
        
        \draw[->] (decision.east) -- node[above] {Yes} (estimation.west);
        \draw[->] (decision.south) -- node[right] {No} (reject.north);
        
        \draw[->] (estimation.east) -- (output.west);

    \end{tikzpicture}%
}}
        \caption{}
        \label{fig:goc_two_node}
    \end{subfigure}

    \caption{\small{Fig.~(a) shows a DC outsourcing the computation of $L(\mathbf{W})$ to worker nodes due to limited resources. Honest workers follow the protocol but may return bounded noisy outputs, while adversaries may strategically inject arbitrary noise. We focus on the \textbf{adversary dominated} regime. Fig.~(b) shows a two-node game-of-coding instance~\cite{GoCJournal, nodehi2026gamevector, GoDSybil, nodehi2025unknown, nodehi2026game}, with one honest and one adversarial node. The honest node returns $\mathbf{Y}_h = L(\mathbf{W})+\mathbf{N}_h$, where $\|\mathbf{N}_h\|_2\leq \Delta$, while the adversarial node returns $\mathbf{Y}_a = L(\mathbf{W})+\mathbf{N}_a$, with $\mathbf{N}_a$ chosen strategically. The DC accepts reports only if $\|\mathbf{Y}_a-\mathbf{Y}_h\|_2\leq \eta\Delta$, and forms an estimate only upon acceptance.
}}
    \vspace{-7pt}
    \label{fig:intro_overview}
\end{figure*}
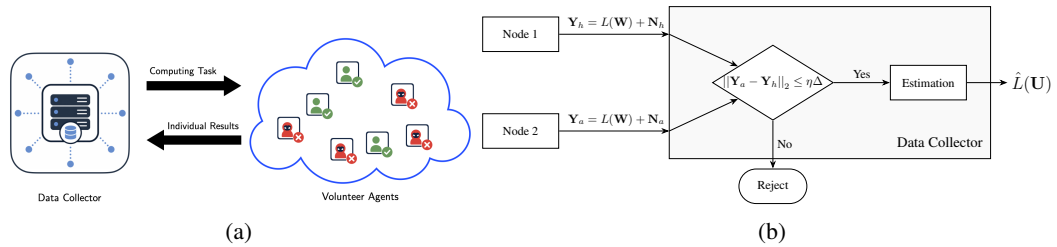

Consider a setting in which a data collector (DC) or an aggregator needs to perform some computational task, but lacks enough  computing power to execute it on its own. To address this limitation, it outsources the task to a set of worker nodes. Because some of these workers may act maliciously, the DC introduces redundancy by assigning the same (or related) computation to multiple nodes. The outsourced task may be a one-time computation, such as model inference, or a sequence of computations, such as minimizing a loss function during training. Some workers are honest and follow the protocol, although their outputs may still contain small inaccuracies due to approximation, quantization, or other benign system effects. Other workers, however, may behave adversarially and deliberately distort the reported computation in order to mislead the DC (see Fig.~\ref{fig:outsourcing_general}).

A key application of this framework is decentralized machine learning (DeML), where smart contracts (a computer program) deployed on a blockchain coordinate training or inference in a transparent and accountable manner~\cite{shafay2023blockchain, ding2022survey, kayikci2024blockchain, taherdoost2023blockchain, taherdoost2022blockchain, tian2022blockchain, salah2019blockchain}. In this context, the smart contract serves as the DC. Due to the  computational limitations of blockchains, AI workloads are delegated to external worker nodes~\cite{zhao2021veriml}. Also, these environments are typically permissionless, meaning that participating nodes may be adversarial, and in fact even the existence of an honest majority cannot be guaranteed~\cite{sliwinski2019blockchains, han2021fact, gans2023zero}.

When the environment is dominated by honest nodes, the DC can rely on robust aggregation methods, such as the geometric median, Krum, coordinate-wise median, to cope with the effect of adversarial behavior~\cite{allen2020byzantine, farhadkhani2022byzantine, shi2025optimal, karimireddy2021learning, guerraoui2024byzantine, zhu2023byzantine, liu2023byzantine, yang2019byzantine, rajput2019detox, li2021byzantine, allouah2023robust, alistarh2018byzantine, el2021collaborative, guerraoui2018hidden, yin2018byzantine, blanchard2017machine}. However, these methods generally lose their effectiveness when honest nodes are in the minority. To protect the computation in \textbf{adversary dominated} settings, the DC imposes a strict acceptance rule: the reported results are accepted only if the values returned by different workers are sufficiently close to one another, within a prescribed tolerance. Since at least a fraction, even one, of the workers are honest, the acceptance rule provides a basic level of reliability by rejecting highly divergent submissions. To motivate the adversarial nodes to collaborate, the DC also employs a reward policy under which all participating nodes, including both honest and adversarial ones, are compensated once the reported values satisfy the acceptance rule. For example, in DeML applications, this reward may take the form of cryptocurrency or token. This incentive mechanism fundamentally changes the role of the adversary. Rather than behaving as a pure saboteur, the adversary becomes a rational agent with a utility function that depends on two competing objectives: increasing the estimation error seen by the DC, and maintaining a sufficiently high probability of acceptance in order to receive the reward. As a result, the adversary must strategically choose its noise-injection. 

Outsourced computation tasks generally fall into two paradigms: single-shot operations, such as model inference, and multi-shot operations, such as iterative gradient descent. In the single-shot setting, the DC also faces a fundamental tradeoff. It prefers the computation to be accepted, since acceptance makes the result available and therefore encourages a more permissive acceptance rule, even at the risk of admitting adversarial outputs. Also, it seeks to keep the estimation error as small as possible, which favors a stricter acceptance rule. To balance these competing objectives, the DC relies on its own utility function to strategically calibrate the acceptance rule. This interaction naturally leads to a  leader-follower game-theoretic framework: first, the DC, as the leader, announces the acceptance rule, and then the adversary, as the follower, chooses its noise  in response (see Fig.~\ref{fig:goc_two_node}). This specific single-round dynamic, particularly in networks dominated by adversary, has been formalized and solved under the game of coding framework~\cite{GoCJournal, nodehi2026gamevector, GoDSybil, nodehi2025unknown, nodehi2026game}, which we detail in Section \ref{sec:Related Works}.

The dynamics shift significantly in multi-shot computational tasks, such as iterative optimization. Unlike single-shot scenarios, the DC does not maintain myopic, instantaneous preferences for individual computation rounds. Instead, the DC is driven by long-term objectives encompassing the entire sequence of interactions. In gradient descent, for example, the DC’s priority is not the marginal outcome of a single iteration, but rather the guarantee of rapid global convergence or the attainment of a specific accuracy threshold relative to the optimum. Consequently, a critical challenge arises: how can the DC’s long-term objectives be translated into a sequence of per-round acceptance rules that ensure a reliable convergence rate? This paper aims to bridge this gap.

\begin{figure}[t]
    \centering
    \includegraphics[width=0.7\linewidth]{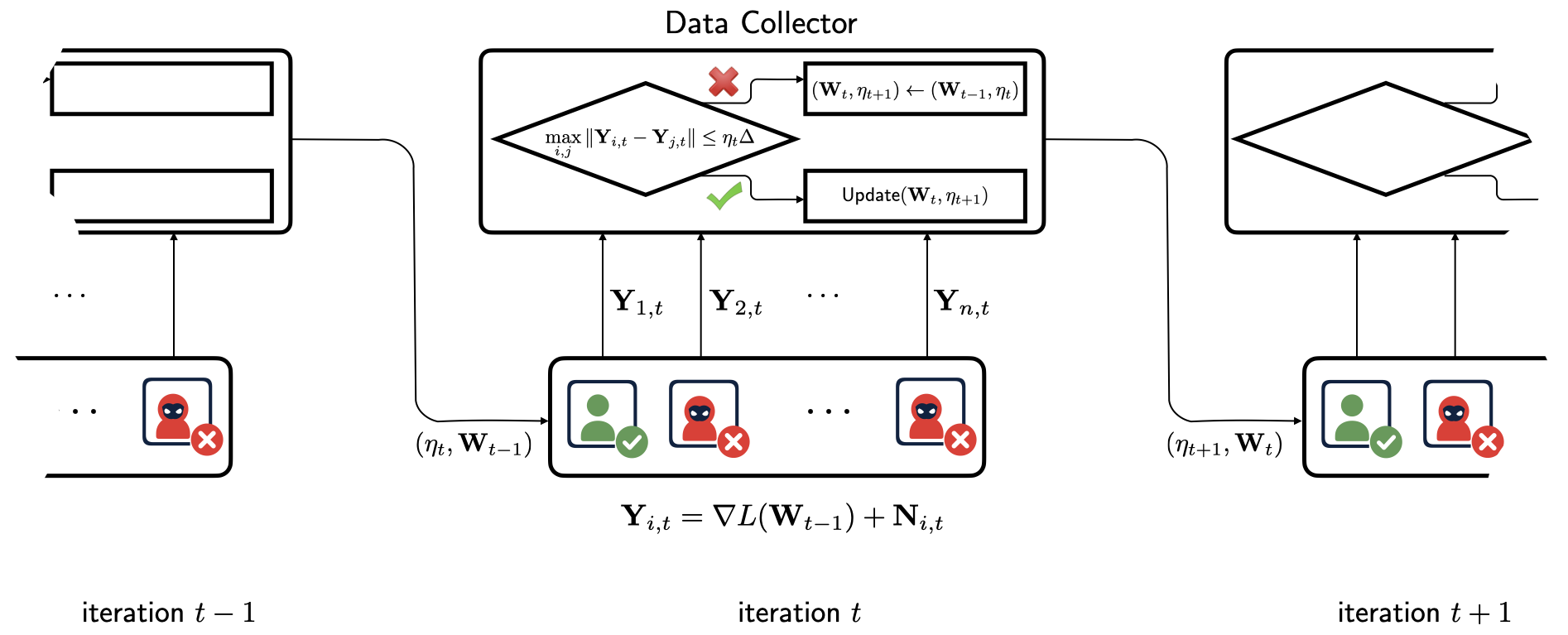}
    \caption{\small{Iterative game of coding framework. Since the DC does not have sufficient resources to evaluate $\nabla L(\mathbf{W}_{t-1})$ on its own, at each round $t$ it broadcasts the current model parameter $\mathbf{W}_{t-1}$ and the acceptance parameter $\eta_t$ to the worker nodes. Each node $i$ returns a noisy gradient report $\mathbf{Y}_{i,t}=\nabla L(\mathbf{W}_{t-1})+\mathbf{N}_{i,t}$. For honest workers, we have $\|\mathbf{N}_{i,t}\|_2\leq \Delta$, while adversarial workers choose their noise strategically according to their utility function. The DC accepts the reports only if $\|\mathbf{Y}_{u,t}-\mathbf{Y}_{v,t}\|_2\leq \eta_t\Delta$ for all pairs of nodes $u,v$. If accepted, it forms an estimate $\widehat{\mathbf{G}}_t$ and updates the model as $\mathbf{W}_{t}\leftarrow \mathbf{W}_{t-1}-\lrtrandom \widehat{\mathbf{G}}_t$; otherwise, the model remains unchanged, that is, $\mathbf{W}_{t}\leftarrow \mathbf{W}_{t-1}$.}}
    \vspace{-20pt}\label{fig:iterative_game_timeline}
\end{figure}

To be specific, in this paper, we consider the problem of minimizing a function $L : \mathbb{R}^{\ndim} \to \mathbb{R}$, which represents the optimization objective of interest. At each round (or equivalently iteration) $t \in \{0,1,2,\dots\}$, the objective is to update the current model parameter $\mathbf{W}_{t-1} \in \mathbb{R}^{\ndim}$  as ${\mathbf{W}_{t}\leftarrow  \mathbf{W}_{t-1} - \lrtrandom \nabla L(\mathbf{W}_{t-1})}$, where $\lrtrandom$ is the learning rate. However, the DC does not have computation power to compute $\nabla L(\mathbf{W}_{t-1})$. As an alternative, the DC announces the current model parameter $\mathbf{W}_{t-1}$ together with an acceptance parameter $\eta_t$ to all volunteer worker nodes. Each node $i$ returns a noisy gradient estimate ${\mathbf{Y}_{i,t} = \nabla L(\mathbf{W}_{t-1}) + \mathbf{N}_{i,t}}$. For honest workers,  the noise $\mathbf{N}_{i,t}$ represents inherent system perturbations, such as quantization or approximation errors, and is typically bounded by $\| \mathbf{N}_{i,t}\|_2 \leq  \Delta$.  For adversarial nodes, $\mathbf{N}_{i,t}$ is sampled from a distribution strategically designed (as a function of $\eta_t$) to balance its incentive for reward (i.e., having its computation accepted) with its objective of disrupting the optimization process (i.e., increasing the estimation error). Upon receiving these reports, the DC applies an acceptance policy requiring that $\|\mathbf{Y}_{u,t} - \mathbf{Y}_{v,t}\|_2 \leq \eta_t \Delta$ for all pairs of nodes  $u$ and $v$. As noted earlier, because at least a fraction of the workers are honest and produce similar results, this condition provides a baseline level of reliability (see Fig.~\ref{fig:iterative_game_timeline}).

Importantly, this acceptance verification is inherently a probabilistic event. Indeed,  the condition $\|\mathbf{Y}_{u,t} - \mathbf{Y}_{v,t}\|_2 \leq \eta_t \Delta$ is equivalent to $\|\mathbf{N}_{u,t} - \mathbf{N}_{v,t}\|_2 \leq \eta_t \Delta$. Thus, even if the adversary strategically selects its injected noise, it does not know the realization of the honest noise in advance. As a result, it cannot guarantee acceptance, and instead must balance the amount of distortion it introduces against the probability that its report will still be accepted. Whenever the computation is accepted, the DC forms an estimate $\widehat{\mathbf{G}}_t$ of the gradient $\nabla L(\mathbf{W}_{t-1})$. This estimate is generally noisy, reflecting both the benign randomness of honest computation and the strategic perturbations introduced by adversarial nodes, as well as the  stochastic variance inherent in algorithms such as stochastic gradient descent. The DC then updates the model according to $\mathbf{W}_{t} = \mathbf{W}_{t-1} - \lrtrandom \widehat{\mathbf{G}}_t$. If the computation is rejected, then no update is performed, that is, $\mathbf{W}_{t} = \mathbf{W}_{t-1}$.

The DC's objective is to design the sequence of acceptance parameters $\{\eta_t\}_{t=0}^\infty$ and learning rates $\{\lrtrandom\}_{t=0}^\infty$ to guarantee convergence over time, in this \textbf{adversary dominated} network. In each iteration $t$ for choosing $\eta_t$, the DC faces a dilemma: 

\begin{mathbox}
{\bf DC Dilemma:} Choosing a larger $\eta_t$ relaxes the acceptance condition, increasing the likelihood that $\mathbf{W}_{t-1}$ is updated, but at the cost of higher error in estimating $\nabla L(\mathbf{W}_{t-1})$. Conversely, choosing a smaller $\eta_t$ tightens the acceptance condition, reducing the likelihood that updates to $\mathbf{W}_{t-1}$ are accepted; however, when updates do occur, the estimation error in $\nabla L(\mathbf{W}_{t-1})$ is smaller on average.
\end{mathbox}

This paper resolves this dilemma by proposing an algorithm that adaptively selects the acceptance parameters $\{\eta_t\}_{t=0}^\infty$ and learning rates $\{\lrtrandom\}_{t=0}^\infty$, achieving near-optimal convergence rates over time, even under adversary dominated network conditions.
At each iteration $t$, the proposed algorithm, called \longalgoname ~(\algoname), 
dynamically determines $\eta_t$ and learning rate $b_t$, based on the history of the outcomes. 

\begin{figure}[t]
    \centering

    \begin{subfigure}[t]{0.49\textwidth}
        \centering
        \includegraphics[width=\linewidth]{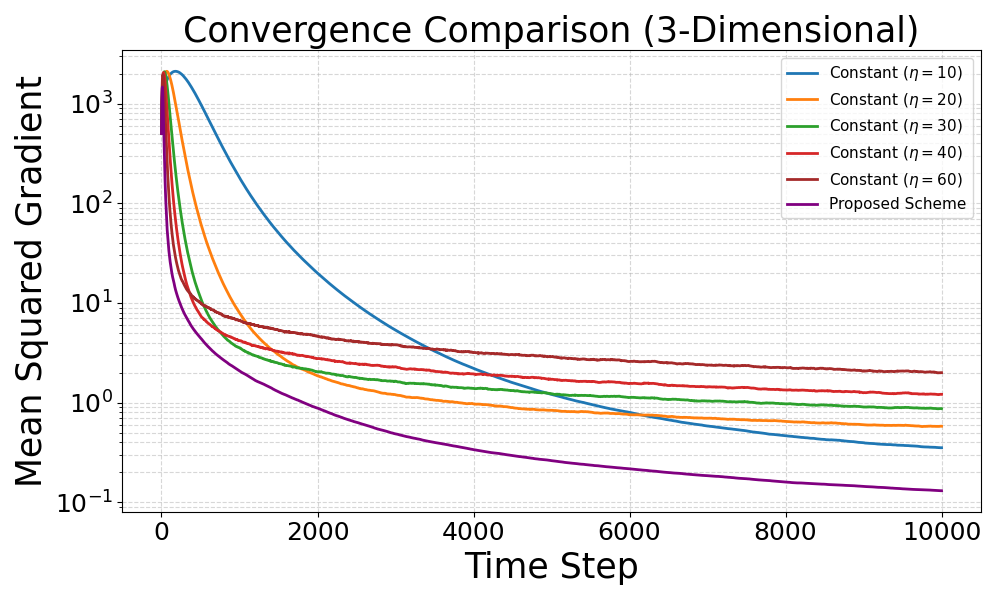}
        \caption{A three-dimensional function}
        \label{fig:experimental_results}
    \end{subfigure}
    \hfill
    \begin{subfigure}[t]{0.44\textwidth}
        \centering
        \includegraphics[width=\linewidth]{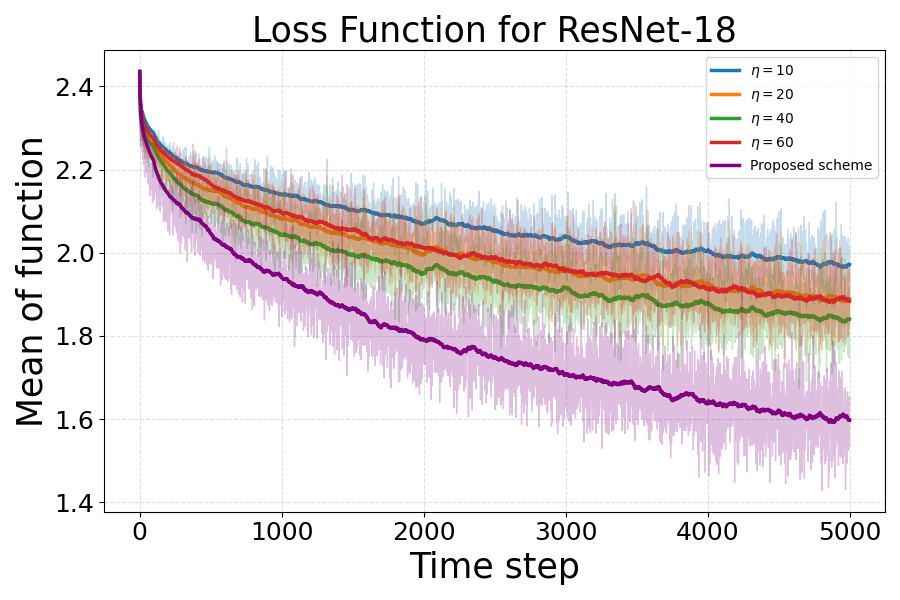}
   \caption{\small{Training loss on CIFAR-10 for ResNet-18.}}
        \label{fig:resnet_loss_results}
    \end{subfigure}

    \caption{\small{ 
    A larger constant $\eta$ enables faster initial progress through more frequent updates, but leads to higher error by admitting more adversarial noise. Smaller ones are more reliable but progress slowly due to frequent rejections. \algoname~adapts $\eta_t$ over time, using larger $\eta_t$ early and smaller  later to achieve best performance.
}}
    \label{fig:side_by_side}
    \vspace{-5mm}
\end{figure}

Our main contributions are twofold:

\textbf{1. Adaptive Design and Empirical Validation:} We empirically demonstrate that \algoname\ significantly outperforms baseline schemes that rely on a constant $\eta$. As illustrated in Fig.~\ref{fig:experimental_results}, the static baselines exhibit a clear ``Tortoise-and-Hare effect'', which highlights the fundamental tradeoff between update frequency and update quality. For example, a highly permissive threshold, such as the brown curve with $\eta = 60$, allows frequent updates and yields rapid initial progress, but eventually plateaus at a relatively high error floor because it admits substantial adversarial noise (like the hare). In this phase, the norm of the gradient decays very slowly.  In contrast, a stricter threshold, such as the blue curve with $\eta = 10$, suffers from a much slower start due to frequent computation rejections, yet eventually overtakes the more permissive baselines and achieves a considerably lower final error (like the tortoise). This behavior shows that any static choice of $\eta$ is fundamentally inadequate, since the optimal threshold changes over the course of the optimization. More precisely, at the beginning, the model is far from the optimum, so the true gradient is large and dominates the adversarial perturbation. In this regime, a larger $\eta$ is desirable because it enables more frequent updates and faster progress toward the optimum. Later, as the model approaches the optimum, the true gradient becomes small, and the adversarial noise can become comparable to, or even dominate, the useful signal. In that regime, $\eta$ must be reduced in order to enforce more accurate and less corrupted updates. Our theoretical analysis of \emph{the expected descent} in each iteration perfectly captures this phenomenon, and guides us to develop \algoname, which resolves this tension by dynamically adapting $\eta_t$ and $b_t$ over time. As it is shown by the purple curve in Fig.~\ref{fig:experimental_results}, the proposed algorithm outperforms any other choice of constant $\eta$ at every single iteration. 
 The proposed algorithm uses a larger $\eta_t$ in the early stage to accelerate descent, and then progressively tightens the acceptance rule as the model approaches the optimum in order to suppress adversarial noise. As shown in Fig.~\ref{fig:experimental_results}, this dynamic strategy achieves the best of both worlds, delivering both fast initial convergence and low final error. Additional experiments under other settings are provided in Section~\ref{Sec:Experiments}.

\textbf{2. Asymptotic Optimality and Convergence Guarantees:} We prove in Theorem~\ref{thm:convergence} that the asymptotic convergence rate of \algoname\ matches the standard convergence rates of Stochastic Gradient Descent (SGD). To the best of our knowledge, this is the first rigorous mathematical framework for decentralized machine learning that guarantees asymptotic rates matching  honest-majority settings, while operating in an adversary dominated network.

\textbf{Notation:}
We use uppercase letters for random variables and lowercase letters for deterministic values. Vectors are denoted in boldface and scalars in standard type. Unless stated otherwise, all vectors lie in $\mathbb{R}^\ndim$, and for $\mathbf{x} = (x_1, \dots, x_\ndim)$, we use the Euclidean norm $\|\mathbf{x}\|_2 = (\sum_{i=1}^\ndim x_i^2)^{1/2}$.

\vspace{-10pt}
\section{Problem Formulation}
\vspace{-10pt}
We consider a multi-round decentralized optimization setting, where the DC aims to minimize an objective function $L : \mathbb{R}^{\ndim} \to \mathbb{R}$. Examples of such objective functions include the training loss of a model over the parameter vector $\mathbf{w} \in \mathbb{R}^{\ndim}$. The DC aims to perform (stochastic) gradient descent, and outsources the calculation of the gradient $\nabla L(\mathbf{w}) \in \mathbb{R}^{\ndim}$ to a network of $\Nnodes$ external worker nodes, denoted by the set $\allnodesset = \{1, \dots, \Nnodes\}$. 
The network is partitioned into two disjoint sets: a set of honest nodes $\honsetset$ and a set of adversarial nodes $\adversaryset$, where $\allnodesset = \honsetset \cup \adversaryset$. We specifically focus on the adversary dominated regime, where $|\adversaryset| \ge |\honsetset|$. The true identities of the nodes are unknown to the DC.

The optimization process unfolds over discrete iterative rounds $t \in \{0, 1, 2, \dots\}$. At each round~$t$, the DC broadcasts the current model parameters $\mathbf{W}_{t-1} \in \mathbb{R}^{\ndim}$ and an acceptance parameter $\eta_t$ to all nodes in $\allnodesset$. For the initial round $t=0$, the DC uses the initial weight vector $\mathbf{W}_{-1}$. Each honest node $i \in \honsetset$ faithfully follows the protocol to calculate the true gradient $\nabla L(\mathbf{W}_{t-1})$. However, due to inherent computational limitations, such as stochastic approximations or floating point inaccuracies, the computed result contains random noise. The report sent to the DC by node $i$ is $\mathbf{Y}_{i,t} = \nabla L(\mathbf{W}_{t-1}) + \mathbf{N}_{i,t}$. For an honest node $i\in \honsetset$, the noise $\mathbf{N}_{i,t} \in \mathbb{R}^{\ndim}$ accounts for various imperfections, such as, randomness of the batch of data used for gradient evaluation, randomness of the data sample in SGD,  or quantization. It is assumed to be bounded $\|\mathbf{N}_{i,t}\|_2 \le \Delta$ with some symmetric distribution. In contrast, the adversarial nodes $j \in \adversaryset$ are assumed to possess the exact realization of the true gradient $\nabla L(\mathbf{W}_{t-1})$. They collaborate to inject malicious noise into the system to corrupt the computation. The report sent by adversarial node $j$ is $\mathbf{Y}_{j,t} = \nabla L(\mathbf{W}_{t-1}) + \mathbf{N}_{j,t}$,
where the injected adversarial noise vectors $\{\mathbf{N}_{j,t}\}_{j \in \adversaryset} \subset \mathbb{R}^{\ndim}$ are drawn from an arbitrary joint probability density function $g(\cdot)$, strategically chosen by the adversary.

Upon receiving $\underline{\mathbf{Y}}_t \triangleq \{\mathbf{Y}_{1,t}, \dots, \mathbf{Y}_{\Nnodes,t}\}=\{\mathbf{Y}_{i,t}:i\in \honsetset\}\cup \{\mathbf{Y}_{j,t}: j\in \adversaryset\}$, the DC first performs a consistency check parameterized by $\eta_t$. More precisely, the \textbf{acceptance event}, denoted by $\acceptancet$, occurs if and only if $\|\mathbf{Y}_{u,t} - \mathbf{Y}_{v,t}\|_2 \le \eta_t \Delta$ for all $u, v \in \allnodesset$. If the computation is accepted, the DC aggregates the reports using a predefined estimation function ${\mathsf{est} : (\mathbb{R}^{\ndim})^{\Nnodes} \to \mathbb{R}^{\ndim}}$ to form an estimated gradient $\widehat{\mathbf{G}}_t \in \mathbb{R}^{\ndim}$. The estimated gradient could be a simple average ${\mathsf{est}(\mathbf{Y}_{1,t}, \dots, \mathbf{Y}_{\Nnodes,t})=\frac{1}{n} \sum_{i=1}^n \mathbf{Y}_{i,t}}$, or any sophisticated estimate. The DC then updates the model parameters using a learning rate $\lrtrandom > 0$ as ${\mathbf{W}_{t} = \mathbf{W}_{t-1} - \lrtrandom \widehat{\mathbf{G}}_t}$. Conversely, if the computation is rejected, the model parameters remain unchanged, that is, $\mathbf{W}_{t} = \mathbf{W}_{t-1}$.
We define the history set $\mathcal{H}_t \triangleq \bigcup_{k=0}^{t}
\{\mathbf{Y}_{i,k}\}_{i\in\mathcal V}$.
Moreover, the accepted gradient estimator is assumed to be conditionally unbiased, meaning that $\mathbb{E}[\widehat{\mathbf{G}}_t \mid \mathcal{H}_{t-1}, \acceptancet ] = \nabla L(\mathbf{W}_{t-1})$ (see Remark~\ref{remak:bias}). Since honesty-based reports with error $\Delta$ can diverge by up to $2\Delta$, we assume $\eta_t \ge 2$ to ensure the DC does not reject valid computations.

The interaction between the DC and the adversary fundamentally depends on two statistical metrics governed by the choice of $\eta_t$ and the adversary's chosen noise distribution $g(\cdot)$. Crucially, the acceptance event is inherently probabilistic. Although the adversary can strategically craft its own noise distribution $g(\cdot)$, it does not know the exact realization of the inherent random noise generated by the honest nodes. Because the adversarial nodes operate without knowing the honest nodes' exact values, they cannot deterministically guarantee the acceptance of their reported results $\{\mathbf{Y}_{i,t}\}_{i\in \adversaryset}$. Instead, the success of the threshold test depends entirely on how the adversary's chosen noise distribution interacts with the acceptance parameter $\eta_t$. We formalize this by defining the probability of acceptance, which represents the system's liveness at round $t$, as $\pa(g(\cdot), \eta_t) \triangleq \Pr(\acceptancet \mid \mathcal{H}_{t-1})$,
where the probability is evaluated over the fresh randomness of the current round, conditioned on the realized history. Furthermore, if the reports are accepted, we evaluate the resulting estimate using the mean squared error $\mse(g(\cdot), \eta_t) \triangleq \mathbb{E} [ \|\nabla L(\mathbf{W}_{t-1}) - \widehat{\mathbf{G}}_t\|_2^2 \mathrel{|} \mathcal{H}_{t-1}, \acceptancet ]$.

The adversary faces a fundamental tradeoff between two competing objectives. On one hand, it seeks to maximize the error. On the other hand, since decentralized networks typically reward a node only when its computation is accepted, the adversary is also motivated to maintain a high probability of acceptance,  in order to secure financial rewards. To capture this behavior, we model the interaction as a Stackelberg game~\cite{von2010market}. More precisely, at the beginning of each round $t$, the DC announces its acceptance policy through the  parameter $\eta_t$, and the adversary then chooses its noise distribution $g(\cdot)$ in response. We assume the adversary is \emph{rational} and \emph{myopic}, meaning that at each round it optimizes only its  utility for the current interaction. Accordingly, it selects $g(\cdot)$ to maximize the utility
\begin{align}\label{eq:adv_utility}
    \mathsf{U}_{\mathrm{AD}}(g(\cdot), \eta_t)
    \triangleq
    Q_{\mathrm{AD}}\bigl(\mse(g(\cdot), \eta_t), \pa(g(\cdot), \eta_t)\bigr),
\end{align}
where $Q_{\mathrm{AD}}$ is strictly increasing in both arguments. 
Accordingly, at round $t$, the adversary's optimal response to $\eta_t$ is $g_{\eta_t}^*(\cdot)
    =
    \arg\max
    \mathsf{U}_{\mathrm{AD}}(g(\cdot), \eta_t)$.
This induces the equilibrium acceptance probability and mean squared error at round $t$, denoted by\footnote{The relationship between $\pat$ and $\mset$ has been characterized in other works for both cases where the adversary's utility is known~\cite{GoCJournal, nodehi2026gamevector, GoDSybil} and where it is unknown~\cite{nodehi2025unknown, nodehi2026game}. In this paper, we assume this relationship is given.
}
\begin{equation} \label{eq:PA_t_and_eq:MSE_t}
    \pat \triangleq \pa(g_{\eta_t}^*(\cdot), \eta_t) \quad \text{and} \quad \mset \triangleq \mse(g_{\eta_t}^*(\cdot), \eta_t).
\end{equation}

Unlike the adversary, the DC does not optimize a local, per-round utility function. 
Its objective is instead long-term: to ensure that the optimization trajectory 
makes sustained progress on the global loss $L(\mathbf{W})$ and converges, in the 
nonconvex sense, toward stationarity. Thus, the central challenge for the DC is to 
adaptively design the acceptance thresholds $\{\eta_t\}_{t=0}^{\infty}$ and 
learning rates $\{\lrtrandom\}_{t=0}^{\infty}$ so as to guarantee fast and robust 
convergence despite adversarially corrupted accepted updates.


While learning-rate schedules $\{\lrtrandom\}_{t=0}^{\infty}$ have been extensively studied~\cite{kingma2015adam, zeiler2012adadelta, luo2019adaptive, zaheer2018adaptive, loshchilov2017decoupled, morales2024exponential, polyak1964some, shazeer2018adafactor, cutkosky2019momentum, robbins1951stochastic, bottou2018optimization, duchi2011adaptive, zhuang2020adabelief, reddi2019convergence, ward2020adagrad, sutskever2013importance, chen2018closing, defazio2023learning}, the acceptance sequence $\{\eta_t\}_{t=0}^{\infty}$ introduces a new control dimension. The threshold $\eta_t$ determines how strictly the DC filters reports at each round, and a fixed choice is generally suboptimal. A small static $\eta$ tightly limits the adversary and yields low-error accepted updates, but it also reduces $\pat$ and slows progress. In contrast, a large static $\eta$ increases acceptance and enables frequent updates, but gives the adversary more room to inject harmful noise and increase $\mset$.

The strategy must therefore adapt $\eta_t$ over time. Far from the optimum $\mathbf{W}^*$, the gradient norm $\|\nabla L(\mathbf{W}_t)\|_2$ is large, so the DC can use a larger threshold to increase $\pat$ and obtain more frequent updates, while the gradient signal still dominates the bounded adversarial noise. Near $\mathbf{W}^*$, the gradient norm shrinks, and a large $\eta_t$ allows adversarial noise to dominate the update and hinder convergence. Thus, the DC must gradually reduce $\eta_t$ as the gradient signal weakens. This paper formalizes this principle and characterizes the decay of the acceptance sequence $\{\eta_t\}_{t=0}^{\infty}$, jointly with the learning rate, to ensure robust convergence under strategic, dominant adversaries.

\begin{remark} \label{remak:bias}
    As discussed in~\eqref{eq:adv_utility}, the adversary has a per-round utility function. Under this myopic adversary model, prior game-of-coding works show that the equilibrium noise strategy is symmetric~\cite{GoCJournal,GoDSybil,nodehi2026gamevector}, which leads to an unbiased gradient estimate at the DC. By contrast, injecting a persistent directional bias would correspond to a non-myopic adversary with a multi-round utility. Learning and mechanism design against such non-myopic agents is a challenging problem in game theory~\cite{haghtalab2022learning,collina2024repeated, arunachaleswaran2024pareto, celentani1996maintaining, chen2023learning}, and we leave its study in this setting for future work.
\end{remark}
\vspace{-10pt}

\section{MainResults}
\vspace{-5pt}
In this section, we first motivate and introduce our proposed algorithm, \emph{\longalgoname}~(\algoname).

The update rule at iteration $t$ can be expressed as ${\mathbf{W}_t = \mathbf{W}_{t-1} - \indicatoraceptance_t \lrtrandom \widehat{\mathbf{G}}_t}$, where $\indicatoraceptance_t=\mathbbm{1}\{\acceptancet\}$ indicates whether the reports are accepted in round $t$, and $\lrtrandom$ is the learning rate, which is a function of the history $\mathcal{H}_{t-1}$. Then, using the descent lemma~\cite{nesterov2004introductory} for $\ell$-smooth functions, we can show
\begin{equation} 
\label{eq:int_final_expected_descent}
    \mathbb{E}[L(\mathbf{W}_t) \mid \mathcal{H}_{t-1}] 
    \le 
    L(\mathbf{W}_{t-1}) 
    - \lrtrandom \pat ( 1 - \frac{\ell \lrtrandom}{2} ) \|\nabla L(\mathbf{W}_{t-1})\|_2^2 
    + \frac{\ell \lrtrandom^2}{2} \pat \mset .
\end{equation}
Let us define the \emph{expected descent} at iteration $t$ as $\descentstep \triangleq \mathbb{E}[L(\mathbf{W}_{t-1}) - L(\mathbf{W}_t) \mid \mathcal{H}_{t-1}]$, which ideally should be positive and large. From~\eqref{eq:int_final_expected_descent}, we have ${\descentstep \ge \pat ( \lrtrandom ( 1 - \frac{\ell \lrtrandom}{2} ) \|\nabla L(\mathbf{W}_{t-1})\|_2^2 - \frac{\ell \lrtrandom^2}{2} \mset )}$. One could use this lower bound as a surrogate objective for $\descentstep$ and select the optimal  $\eta_t^*$ by solving
\begin{equation} 
\label{eq:int_optimal_eta_problem}
    \eta_t^* 
    = 
    \arg\max_{\eta \in [\eta_{\min}, \eta_{\max}]}
    ~
    \pat (
    \lrtrandom ( 1 - \frac{\ell \lrtrandom}{2} ) 
    \|\nabla L(\mathbf{W}_{t-1})\|_2^2 
    - \frac{\ell \lrtrandom^2}{2} \mset
    ).
\end{equation}
However, solving~\eqref{eq:int_optimal_eta_problem} is challenging and computationally intensive, since the relationship between $\pat$ and $\mset$ is nonlinear and is induced by the adversary's utility in~\eqref{eq:adv_utility}.

To resolve this challenge, let us take a closer look at the surrogate objective. This objective highlights the interaction between a positive term corresponding to the \emph{signal} $\|\nabla L(\mathbf{W}_{t-1})\|_2^2$, and a negative term corresponding to the \emph{noise power} $\mset$. It provides two key insights. \textbf{(i)} The choice of $\eta_t$ creates a fundamental tradeoff: increasing $\eta_t$ increases $\pat$, which promotes more frequent updates and scales up the expected descent; however, it also increases $\mset$, which can reduce $\descentstep$. This tradeoff indeed represents the \emph{DC dilemma} discussed in the introduction. \textbf{(ii)} In the early stages of optimization, the gradient norm $\|\nabla L(\mathbf{W}_{t-1})\|_2^2$ may be large because the model $\mathbf{W}_{t-1}$ may still be far from a stationary point. Therefore, as long as $\mset$ remains small compared to $\|\nabla L(\mathbf{W}_{t-1})\|_2^2$, the useful signal dominates the noise, and the algorithm can tolerate a larger estimation error. In contrast, as the model approaches a stationary point, the gradient norm becomes small, that is, $\|\nabla L(\mathbf{W}_{t-1})\|_2^2 \approx 0$. In this regime, the noise term becomes comparatively more significant, and $\eta_t$ must be adjusted to limit $\mset$.
This observation suggests choosing $\eta_t$ so that $\mset = c\|\nabla L(\mathbf{W}_{t-1})\|_2^2$ for some hyperparameter $c>0$, thereby keeping the noise level proportional to the current signal strength. The hyperparameter $c$ absorbs the effects of other coefficients in the surrogate objective, such as the Lipschitz constant $\ell$. If the resulting threshold is smaller than $\eta_{\min}=2$, then keeping the noise proportional to the signal is no longer possible within the feasible range, and the DC chooses the strictest admissible threshold $\eta_t=2$, equivalently $\mset=\sigmin$. In addition, we restrict $\eta_t$ to be at most $\eta_{\max}$.
Putting these considerations together, the DC sets $\eta_t$ as the smallest $\eta$ satisfying
\begin{equation}
\label{eq:target_variance_rule_gradient}
    \mse(g_{\eta}^*,\eta)
    =
    \targetvariance
    \triangleq
    \min\{\sigmax,\max\{\sigmin,\, c\|\nabla L(\mathbf{W}_{t-1})\|_2^2\}\}, \qquad \eta \in [\eta_{\min}, \eta_{\max}].
\end{equation}
 Note that \eqref{eq:target_variance_rule_gradient} can be efficiently solved using a binary search since both $\mset\in [\sigma^2_{\min}, \sigma^2_{\max}]$ and $\pat\in[p_{\min}, p_{\max}]$ are an increasing function of $\eta_t$.

Another point that needs to be clarified is the choice of the learning rate $\lrtrandom$. Whenever ${\eta_t > \eta_{\min}}$, which we refer to as the \textit{adaptive regime}, the gradient norm is still large enough to guarantee  
${\mset \leq c\|\nabla L(\mathbf{W}_{t-1})\|_2^2}$, i.e., the noise power is upper bounded by a
constant multiple of the squared gradient norm. This is reminiscent of the strong growth condition~\cite{Schmidt2013}, where the stochastic error is bounded proportionally to the gradient norm. Motivated by this analogy, we keep the learning rate fixed within the adaptive regime, i.e., $\lrtrandom = B_{t-1}$, so that the algorithm can benefit from larger updates while the gradient signal dominates the noise.
When $\eta_t$ reaches $\eta_{\min}$, the threshold cannot be tightened further, and the system enters the \textit{saturated regime}. In this regime, the estimation error is saturated at its minimum achievable level, $\mset=\sigmin$. This is analogous to noisy or stochastic gradient descent with a persistent variance floor, where a decaying learning rate is required to guarantee convergence. Following the standard stochastic optimization literature~\cite{kingma2015adam, zeiler2012adadelta, luo2019adaptive, zaheer2018adaptive, loshchilov2017decoupled, morales2024exponential, polyak1964some, shazeer2018adafactor, cutkosky2019momentum, robbins1951stochastic, bottou2018optimization, duchi2011adaptive, zhuang2020adabelief, reddi2019convergence, ward2020adagrad, sutskever2013importance, chen2018closing, defazio2023learning}, we therefore use a learning rate that decays on the order of $\mathcal{O}(1/\sqrt{t})$ during the saturated regime.

The other less immediate issue is that when the DC selects $\eta_t$ according to \eqref{eq:target_variance_rule_gradient}, it does not yet have access to $\|\nabla L(\mathbf{W}_{t-1})\|_2^2$ or a fresh estimate of it. Indeed, announcing $\eta_t$ is a prerequisite for asking the worker nodes to compute and report $\nabla L(\mathbf{W}_{t-1})$. To address this, for practical implementation, we use a bias-corrected exponential moving average of past accepted gradient estimates of $\{\nabla L(\mathbf{W}_{k})\}_{k=0}^{t-2}$. We denote this one-step proxy for $\nabla L(\mathbf{W}_{t-1})$ by $\widetilde{\mathbf{G}}_{t-1}$.  Nevertheless,  for theoretical analysis, we assume that an oracle provides the gradient norm $\|\nabla L(\mathbf{W}_k)\|_2^2$ to the DC. The algorithm is not highly sensitive to small errors in this prediction, and empirically, this one-step proxy is sufficient for it, as discussed in Section~\ref{Sec:Experiments},
where our experimental results corroborate the convergence of \algoname, under using the proxy, instead of the actual gradient norm. 


Algorithm~\ref{alg:avm_dgd} outlines the execution of \algoname. The algorithm tracks two counters, $\numberofacceptance_t=\sum_{k=0}^{t}\indicatoraceptance_k$ and $\tau_t=\sum_{k=0}^{t}\indicatorsaturate_k$, where $\indicatoraceptance_k=1$ if round $k$ is accepted, and $\indicatorsaturate_k=1$ only if round $k$ is accepted and $\eta_k=\eta_{\min}$. Thus, $\numberofacceptance_t$ is the number of accepted rounds up to round $t$, while $\tau_t$ is the number of accepted saturated rounds.
At each round $t\geq 0$, the DC broadcasts $(\mathbf{W}_{t-1},\eta_t)$ to all workers and receives the reports $\underline{\mathbf{Y}}_t$. If round $t$ is accepted, the DC forms $\widehat{\mathbf{G}}_t=\mathsf{est}(\underline{\mathbf{Y}}_t)$ and updates $\mathbf{W}_{t}=\mathbf{W}_{t-1}-\lrtrandom \widehat{\mathbf{G}}_t$, where $\lrtrandom=b_0(\tau_{t-1}+1)^{-1/2}$. It also updates $\mathbf{M}_t=\beta \mathbf{M}_{t-1}+(1-\beta)\widehat{\mathbf{G}}_t$ and $\widetilde{\mathbf{G}}_t=\mathbf{M}_t/(1-\beta^{\numberofacceptance_t})$. If the computation is rejected, then $\mathbf{W}_t$, $\mathbf{M}_t$, and $\widetilde{\mathbf{G}}_t$ remain unchanged. We use $\|\widetilde{\mathbf{G}}_t\|_2^2$ as a proxy for $\|\nabla L(\mathbf{W}_{t})\|_2^2$.  Therefore, instead of using \eqref{eq:target_variance_rule_gradient}, the DC chooses $\eta_{t+1}$ as the smallest value of $\eta \in [\eta_{\min}, \eta_{\max}]$ satisfying $\mse(g_{\eta}^*,\eta)=\min\{\sigmax,\max\{\sigmin,\, c\|\widetilde{\mathbf{G}}_{t}\|_2^2\}\}$.

We now establish the convergence of \algoname. We assume that $L(\mathbf{w}) \ge L^*$ for all $\mathbf{w} \in \mathbb{R}^{\ndim}$. In addition, we assume that $L(\mathbf{w})$ is $\ell$-smooth, that is, for all $\mathbf{u}, \mathbf{v} \in \mathbb{R}^{\ndim}$, ${\|\nabla L(\mathbf{u})-\nabla L(\mathbf{v})\|_2 \le \ell\|\mathbf{u}-\mathbf{v}\|_2}$. Theorem \ref{thm:convergence} characterizes the asymptotic convergence rate of \algoname. 

Recall that $\|\widetilde{\mathbf{G}}_t\|_2^2$ is a one-step prediction of $\|\nabla L(\mathbf{W}_{t})\|_2^2$ based on past accepted gradient estimates of $\{\nabla L(\mathbf{W}_{k})\}_{k=0}^{t-2}$. 
This is motivated by a long line of stochastic optimization methods that use exponential moving averages to smooth noisy gradient information and track first- or second-order gradient statistics~\cite{kingma2015adam, zeiler2012adadelta, luo2019adaptive, zaheer2018adaptive, loshchilov2017decoupled, morales2024exponential, polyak1964some, shazeer2018adafactor, cutkosky2019momentum}. This principle has also been used in distributed and Byzantine-robust optimization, where momentum-based history helps improve robustness to adversarial behavior~\cite{karimireddy2021learning,farhadkhani2022byzantine}. In the theorem below, we analyze the idealized setting in which the approximation ${\|\widetilde{\mathbf{G}}_t\|_2^2 \approx \|\nabla L(\mathbf{W}_{t})\|_2^2}$ is exact up to oracle access to $\|\nabla L(\mathbf{W}_{t})\|_2^2$. More precisely, instead of Line~\ref{alg:eta_update_line} of Algorithm~\ref{alg:avm_dgd}, the threshold $\eta_t$ is determined using \eqref{eq:target_variance_rule_gradient}.

\begin{theorem} \label{thm:convergence}
Let $L(\mathbf{w})$ be $\ell$-smooth and bounded from below by $L^*$. Under the execution of \algoname, where $\eta_t$ is determined using \eqref{eq:target_variance_rule_gradient} instead of Line~\ref{alg:eta_update_line} of Algorithm~\ref{alg:avm_dgd}, with initial learning rate $b_0 \le \frac{1}{\ell(1+c)}$ for some hyperparameter $c>0$, we have $\min_{0 \le t \le T} \mathbb{E}[\|\nabla L(\mathbf{W}_{t-1})\|_2^2] \le \mathcal{O}( \frac{\ln T}{\sqrt{T}})$.
\end{theorem}

\begin{algorithm}[htbp]
\caption{\longalgoname~(\algoname)}
\label{alg:avm_dgd}
\begin{algorithmic}[1]
\Require $b_0 > 0$, $c > 0$, $\beta \in [0, 1)$, $\mathbf{W}_{-1}$.
\State \textbf{Initialize:} $\tau_{-1} \gets 0$, $\numberofacceptance_{-1} \gets 0$, 
$\mathbf{M}_{-1} \gets \mathbf{0}$, $\widetilde{\mathbf{G}}_{-1} \gets \mathbf{0}$, $\eta_0 \gets (\eta_{\min} + \eta_{\max})/2$

\For{round $t = 0,1,\dots$}
    \State The DC broadcasts $(\mathbf{W}_{t-1}, \eta_t)$ and receives $\underline{\mathbf{Y}}_t$
    
    \If{$\acceptancet$ is satisfied}
        \State $\lrtrandom \gets b_0/\sqrt{\tau_{t-1}+1}$, 
        $\widehat{\mathbf{G}}_t \gets \mathsf{est}(\underline{\mathbf{Y}}_t)$,
        $\mathbf{W}_{t} \gets \mathbf{W}_{t-1} - \lrtrandom \widehat{\mathbf{G}}_t$,
        $\numberofacceptance_{t} \gets \numberofacceptance_{t-1}+1$
        
        \State $\mathbf{M}_{t} \gets \beta \mathbf{M}_{t-1} + (1-\beta)\widehat{\mathbf{G}}_t$,
        $\widetilde{\mathbf{G}}_{t} \gets \mathbf{M}_{t}/(1-\beta^{\numberofacceptance_{t}})$
        
        \State $\eta_{t+1}\gets \inf\{\eta\in[\eta_{\min},\eta_{\max}]:
        \mathsf{MSE}(g^*_\eta,\eta) = \min\{\sigmax,\max\{\sigmin,\, c\|\widetilde{\mathbf{G}}_{t}\|_2^2\}\}\}$ \label{alg:eta_update_line}
        
        \State \textbf{if} $\eta_t = \eta_{\min}$ \textbf{then} $\tau_t \gets \tau_{t-1}+1$ \textbf{else} $\tau_t \gets \tau_{t-1}$
    \Else
        \State $\mathbf{W}_{t} \gets \mathbf{W}_{t-1}$,
        $\mathbf{M}_{t} \gets \mathbf{M}_{t-1}$,
        $\widetilde{\mathbf{G}}_{t} \gets \widetilde{\mathbf{G}}_{t-1}$,
        $\numberofacceptance_{t} \gets \numberofacceptance_{t-1}$,
        $\tau_t \gets \tau_{t-1}$,
        $\eta_{t+1} \gets \eta_t$
    \EndIf
\EndFor
\end{algorithmic}
\end{algorithm}

\vspace{-3pt}
\begin{remark}
The bound in Theorem \ref{thm:convergence}, implies that \algoname\   asymptotically achieves the same convergence rate as standard SGD despite the presence of dominated strategic adversaries.
\end{remark}
\vspace{-5pt}
The detailed proof of Theorem~\ref{thm:convergence} is provided in Appendix~\ref{app:proof_theorem_v2}; here, we highlight the key step. In the appendix, we show that for both the adaptive and saturated regimes, we have
\begin{equation} \label{eq:int_bounded_descent}
    \mathbb{E}[L(\mathbf{W}_t) \mid \mathcal{H}_{t-1}] \le L(\mathbf{W}_{t-1}) - \frac{\lrtrandom \pmin}{2} \|\nabla L(\mathbf{W}_{t-1})\|_2^2 + \frac{\ell \sigmin}{2} \lrtrandom^2 \mathbb{E}[\indicatorsaturate_t \mid \mathcal{H}_{t-1}].
\end{equation}
The last term in \eqref{eq:int_bounded_descent} contributes only in the saturated regime, where $\eta_t=\eta_{\min}$. In the adaptive regime, where $\eta_t>\eta_{\min}$, we have $\indicatorsaturate_t=0$, and hence the noise penalty vanishes. By rearranging \eqref{eq:int_bounded_descent}, taking total expectation, and summing over $t=0,\dots,T$, we obtain
\begin{align} \label{eq:int_summed_descent}
    \frac{\pmin}{2} \sum\nolimits_{t=0}^{T} \mathbb{E}[\lrtrandom \|\nabla L(\mathbf{W}_{t-1})\|_2^2]
    &\le \sum\nolimits_{t=0}^{T} \mathbb{E}\!\left[L(\mathbf{W}_{t-1})-L(\mathbf{W}_t)\right]
    + \frac{\ell \sigmin}{2}\sum\nolimits_{t=0}^{T}\mathbb{E}[\indicatorsaturate_t \lrtrandom^2]
    \nonumber \\
    &\overset{(a)}{\le} 
    L(\mathbf{W}_{-1})-L^* 
    + \frac{\ell \sigmin}{2}\sum\nolimits_{t=0}^{T}\mathbb{E}[\indicatorsaturate_t \lrtrandom^2]
    \overset{(b)}{\le}
    \mathcal{O}(\ln T),
\end{align}
where $(a)$ follows from the telescoping sum
$\sum_{t=0}^{T}(\mathbb{E}[L(\mathbf{W}_{t-1})]-\mathbb{E}[L(\mathbf{W}_t)])=L(\mathbf{W}_{-1})-\mathbb{E}[L(\mathbf{W}_T)]\le L(\mathbf{W}_{-1})-L^*$ . Step $(b)$ follows because the learning rate decays only during accepted saturated rounds. Specifically, if $S\triangleq \sum_{t=0}^{T}\indicatorsaturate_t\le T+1$, then pathwise
$\sum_{t=0}^{T}\indicatorsaturate_t \lrtrandom^2 = b_0^2\sum_{k=1}^{S}\frac{1}{k}\le b_0^2(1+\ln S)\le b_0^2(1+\ln(T+1))$.
This is why eliminating the noise penalty during the adaptive regime is essential. More precisely, in adaptive regime, the learning rate does not decay; therefore, if the penalty $\frac{\ell\sigmin}{2}\lrtrandom^2$ appeared in every round, summing it over $T$ rounds could produce a linear $\mathcal{O}(T)$ term. Instead, \eqref{eq:int_bounded_descent} multiplies this penalty by $\indicatorsaturate_t$, so it contributes only during accepted saturated rounds, where the learning rate decays and the total contribution is only $\mathcal{O}(\ln T)$.

To complete the proof of Theorem \ref{thm:convergence}, we show that
\begin{align} \label{eq:int_weighted_average_trick}
    \min_{0 \le t \le T} \mathbb{E}[\|\nabla L(\mathbf{W}_{t-1})\|_2^2]
    \overset{(a)}{\le}
    \frac{\sum_{t=0}^{T} \mathbb{E}[\lrtrandom \|\nabla L(\mathbf{W}_{t-1})\|_2^2]}
    {b_0\sum_{t=0}^{T}(t+1)^{-1/2}}
    \overset{(b)}{\le}
    \mathcal{O}(\frac{\ln T}{\sqrt{T}}),
\end{align}
where $(a)$ follows since $\tau_{t-1}\le t$, and hence
$\lrtrandom=b_0/\sqrt{\tau_{t-1}+1}\ge b_0/\sqrt{t+1}$, together with the fact that the minimum of a nonnegative sequence is upper bounded by its weighted average. Moreover, $(b)$ follows from \eqref{eq:int_summed_descent} and
$b_0\sum_{t=0}^{T}(t+1)^{-1/2}\ge 2b_0(\sqrt{T+2}-1)=\Omega(\sqrt{T})$.
\vspace{-8pt}

\section{Experimental Results}
\label{Sec:Experiments}
\vspace{-8pt}

In this section, we empirically evaluate \algoname. We use prior works on game of coding~\cite{GoCJournal, nodehi2026gamevector, GoDSybil, nodehi2025unknown, nodehi2026game} to characterize $\pat$, $\mset$, and the estimator $\mathsf{est}(\cdot)$. Unless stated otherwise, we consider $\Nnodes=2$ nodes with $|\honsetset|=|\adversaryset|=1$, and use adversarial utilities of the form $\mathsf{U}_{\mathrm{AD}}=\log(\mse)+\lambda\log(\pa)$. All implementation details are provided in Appendix~\ref{app:exp_details}. We also provide an ablation study on the hyperparameter $c$ in Appendix~\ref{app:c_sensitivity}. For all baselines, ``constant $\eta$'' means that $\eta_t$ is fixed, and $B_t=b_0/\sqrt{\numberofacceptance_{t-1}+1}$, where $\numberofacceptance_{t-1}=\sum_{k=0}^{t-1}\indicatoraceptance_k$. Since rejected rounds do not update the iterate, this gives the same SGD-type decay over accepted rounds, so the comparison isolates the effect of dynamic thresholding. In figures with shaded regions, the shaded area represents standard deviation across independent runs.


We start with the three-dimensional objective $L(\mathbf{w}) = 10w_1\sin(w_2/10) + 10w_2\sin(w_3/10) + 10w_3\sin(w_1/2)$. We set $(\lambda,c,b_0)=(0.03,1,0.1)$. As shown in Fig.~\ref{fig:experimental_results}, smaller constant $\eta$ are more conservative, larger values provide faster initial progress, and \algoname\ achieves the best performance. 

\begin{figure}
\vspace{-3mm}
\centering
\includegraphics[width=0.75\linewidth]{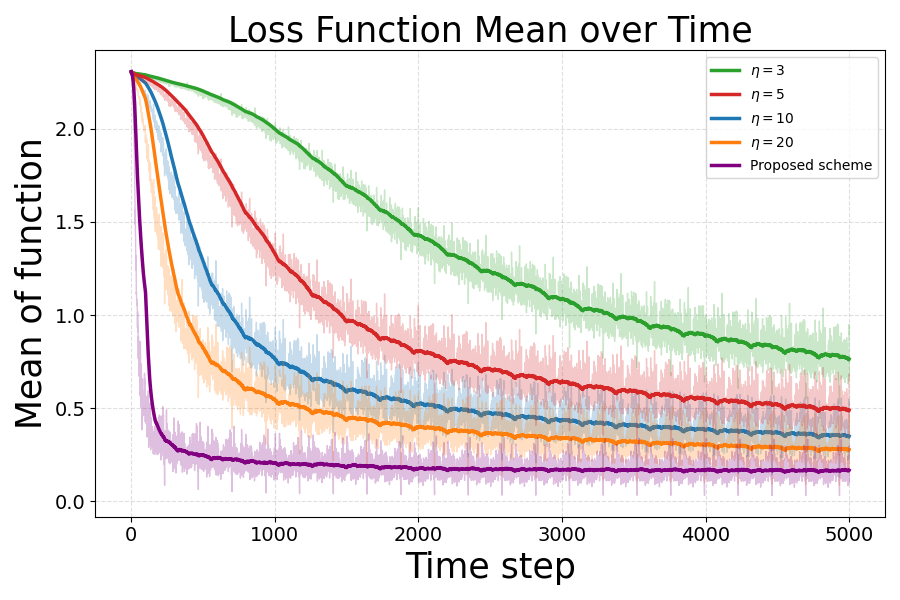}
\vspace{-2mm}
        \caption{\small{Training loss on MNIST for LeNet. }}
        \label{fig:mnist_loss_results}
        \vspace{-6mm}
\end{figure}

Then, we evaluate \algoname\ on a practical learning task by training a LeNet model~\cite{lecun2002gradient} on the MNIST dataset~\cite{lecun2010mnist}, where $\ndim=23942$. We set $(\lambda,c,b_0)=(0.0003,15,0.25)$. We report the average training loss using a moving average window of $100$ for visualization. Fig.~\ref{fig:mnist_loss_results} shows the same pattern: larger values of $\eta$ yield faster initial descent, while smaller values exhibit stronger later-stage improvement. \algoname~ exploits this tradeoff by starting with a relatively large threshold and  reducing it over time. 

Finally, we evaluate \algoname\ by training a ResNet-18 model~\cite{he2016deep} on the CIFAR-10 dataset~\cite{Krizhevsky09learningmultiple}, where the dimension of $\mathbf{w}$ is $\ndim=11173962$. We set $(\lambda,c,b_0)=(0.00001,5,0.01)$. Fig.~\ref{fig:resnet_loss_results} shows a similar pattern, and the proposed scheme again achieves the best overall performance. 

\vspace{-10pt}
\section{Related Works}\label{sec:Related Works}
\vspace{-5pt}

Outsourcing computational tasks to external worker nodes has been widely studied, with federated learning as a prominent example, where a server aggregates local worker updates into a global model~\cite{mcmahan2017communication}. Robust federated learning methods protect this aggregation against adversarial behavior. Examples include Krum, coordinate-wise median, trimmed mean, Bulyan, and related robust SGD variants~\cite{allen2020byzantine, farhadkhani2022byzantine, shi2025optimal, karimireddy2021learning, guerraoui2024byzantine, zhu2023byzantine, liu2023byzantine, yang2019byzantine, rajput2019detox, li2021byzantine, allouah2023robust, alistarh2018byzantine, el2021collaborative, blanchard2017machine, yin2018byzantine, guerraoui2018hidden}. More recent methods further adapt their aggregation rules during training, for example by assigning trust weights, filtering malicious updates, or updating worker reliability scores over time~\cite{cao2021fltrust, nguyen2022flame, mu2024feddmc, huang2025fedid, penalva2025repunet, younesi2025flare}. However, all of these works operate under the honest-majority setting.

In contrast, the focus of this paper is on adversary dominated settings, where $|\adversaryset| \ge |\honsetset|$.
This regime is especially relevant in Web 3.0 systems. Since the emergence of cryptocurrencies and smart contract platforms~\cite{bitcoin2008bitcoin, buterin2013ethereum}, decentralized networks have been designed to operate without trusted intermediaries. A central challenge in this setting is DeML, where heavy computational tasks must be offloaded to off-chain worker nodes because blockchains cannot natively support the computational scale of modern AI~\cite{zhao2021veriml}. In such permissionless environments, these workers are anonymous and untrusted~\cite{sliwinski2019blockchains, han2021fact, gans2023zero}, which leads to a fundamental verification problem: how can the DC trust an outsourced computation without recomputing it?  For DeML applications, a practical outsourcing framework should simultaneously provide low computational overhead, support approximate computation, fast finality, and resilience even under adversarial dominance. 

Existing approaches do not satisfy all of these requirements at once.
Verifiable computing~\cite{thaler2022proofs, feng2021zen} provides strong correctness  by requiring cryptographic proofs, but proof generation remains expensive and slow~\cite{liu2021zkcnn, xing2023zero, mohassel2017secureml, lee2024vcnn, weng2021mystique}, and it is not naturally aligned with approximate computation~\cite{weng2021mystique, chen2022interactive, garg2022succinct, setty2012taking, garg2023experimenting}. Optimistic verification~\cite{bhat2023sakshi, conway2024opml} is more efficient, but delays finality and also struggles with approximate computation. Coded computing~\cite{SudanBook, ZamirCoded, roth2020analog, jahani2018codedsketch,BACC, yu2017polynomial, yu2019lagrange} is fast and naturally compatible with approximate settings, but its guarantees collapse once adversaries become dominant.

This limitation motivated the development of the game of coding framework~\cite{GoCJournal, nodehi2026gamevector, GoDSybil, nodehi2025unknown, nodehi2026game}, which models adversaries as rational, reward seeking agents rather than pure saboteurs. Under this perspective, the framework retains the efficiency of coding-based approaches while remaining effective even when $|\adversaryset| \geq |\honsetset|$. The foundational work~\cite{GoCJournal} introduced the game theoretic formulation and characterized the equilibrium behavior of the participants, showing that reliable operation is possible under adversarial dominance. Subsequent works established Sybil resilience~\cite{GoDSybil}, developed bandit-based methods for learning system parameters when the adversary is unknown~\cite{nodehi2025unknown}, and extended the equilibrium analysis to the general high dimensional setting~\cite{nodehi2026gamevector}.

Despite these advances, prior works on game of coding have focused on single-round computation, where the DC chooses one acceptance rule for one outsourced task. They do not address iterative settings such as multi-round optimization, where the threshold chosen in one round affects not only the current estimate but also the future optimization trajectory through $\mathbf{W}_t$, the gradient norm, and the learning-rate schedule. This paper fills that gap by studying the multi-round setting, characterizing its fundamental tradeoffs, and developing an algorithm for training in adversary dominated environments.
\vspace{-5mm}
\section{Conclusion}

We studied decentralized learning in adversary-dominated settings where honest-majority assumptions may not hold. By combining consistency-based acceptance with adaptive threshold selection, \algoname\ incentivizes workers to submit mutually reliable reports while maintaining progress in iterative optimization. Our theoretical analysis and experiments show that incentive-aware adaptation can recover the asymptotic convergence behavior of standard SGD, even when rational adversaries control a majority of workers. These results suggest that carefully designed acceptance-and-reward mechanisms offer a promising path toward robust permissionless machine learning.

\section{Acknowledgment}
The work of Mohammad Ali Maddah-Ali, Hanzaleh Akbari Nodehi, and Parsa Moradi has been partially supported by the National Science Foundation under Grant CCF-2348638. The work of Soheil Mohajer is  supported in part by AFOSR under the grant FA9550-23-1-0057.

\appendix

\section{Proof of Theorem \ref{thm:convergence}}
\label{app:proof_theorem_v2}
In this section, we present the detailed proof of Theorem~\ref{thm:convergence}. Recall that Theorem~\ref{thm:convergence} analyzes the idealized execution of \algoname, where $\eta_t$ is determined using \eqref{eq:target_variance_rule_gradient} instead of Line~\ref{alg:eta_update_line} of Algorithm~\ref{alg:avm_dgd}. In this setting, the threshold is chosen using oracle access\footnote{Note that the oracle only needs to provide the gradient norm $\|\nabla L(\mathbf{W}_{t-1})\|_2^2$, not the gradient $\nabla L(\mathbf{W}_{t-1})$ itself. It is worth noting that while the latter is a $d$-dimensional vector, its norm is a scalar. } to the gradient norm $\|\nabla L(\mathbf{W}_{t-1})\|_2^2$.

We analyze the convergence of \algoname\ for a non-convex, $\ell$-smooth objective function $L(\mathbf{w})$ that is bounded from below by $L^*$. By the descent lemma \cite{nesterov2004introductory}, the $\ell$-smoothness of $L$ implies that
\begin{equation}
\label{eq:smoothness_app_v2}
L(\mathbf{W}_{t})
\le
L(\mathbf{W}_{t-1})
+
\left\langle \nabla L(\mathbf{W}_{t-1}), \mathbf{W}_{t} - \mathbf{W}_{t-1} \right\rangle
+
\frac{\ell}{2}\|\mathbf{W}_{t} - \mathbf{W}_{t-1}\|_2^2.
\end{equation}
According to the protocol of \algoname\ described in Algorithm \ref{alg:avm_dgd}, the parameter update can be compactly written using an indicator variable $\indicatoraceptance_t \in \{0, 1\}$. Recall that we define $\indicatoraceptance_t = 1$ if the probabilistic acceptance event $\acceptancet$ occurs at round $t$, and $\indicatoraceptance_t = 0$ otherwise.  Thus, the unified parameter update rule can be expressed as
\begin{equation}\label{eq:unified_update_v2}
    \mathbf{W}_{t} = \mathbf{W}_{t-1} - \indicatoraceptance_t \lrtrandom \widehat{\mathbf{G}}_t.
\end{equation}
This formulation is universally applicable; if the computation is rejected ($\indicatoraceptance_t = 0$), the indicator zeroes out the subtracting term, ensuring the parameters remain unchanged ($\mathbf{W}_{t} = \mathbf{W}_{t-1}$). Substituting the update step from \eqref{eq:unified_update_v2} into \eqref{eq:smoothness_app_v2} yields
\begin{equation} \label{eq:descent_step_v2}
    L(\mathbf{W}_{t}) \le L(\mathbf{W}_{t-1}) - \indicatoraceptance_t \lrtrandom \langle \nabla L(\mathbf{W}_{t-1}), \widehat{\mathbf{G}}_t \rangle + \frac{\ell}{2} \indicatoraceptance_t \lrtrandom^2 \|\widehat{\mathbf{G}}_t\|_2^2.
\end{equation}
Recall that we defined history set $\mathcal{H}_t$ as the collection of the realizations of all random variables observed up to round $t$ as
\begin{align}
    \mathcal{H}_t \triangleq \bigcup_{k=0}^{t}
\{\mathbf{Y}_{i,k}\}_{i\in\mathcal V}.
\end{align}
Based on this definition and the protocol in Algorithm~\ref{alg:avm_dgd}, note that the learning rates $\{B_k\}_{k=0}^{t}$ are uniquely determined by the history $\mathcal{H}_{t-1}$. In addition, based on Line~\ref{alg:eta_update_line} of Algorithm~\ref{alg:avm_dgd}, $\eta_t$ is also uniquely determined by the history $\mathcal{H}_{t-1}$. As a consequence, the learning rate $\lrtrandom$ and the threshold $\eta_t$ are random variables that are deterministically evaluated once the history $\mathcal{H}_{t-1}$ is realized. 
Therefore, when taking the expectation of \eqref{eq:descent_step_v2} conditioned on $\mathcal{H}_{t-1}$, the learning rate $\lrtrandom$ can be treated as a constant factor and moved outside the expectation. Moreover, conditioned on $\mathcal{H}_{t-1}$, the random variable  $\mathbf{W}_{t-1}$ is also deterministic. This yields to
\begin{align} \label{eq:exp_descent_step_v2}
    \mathbb{E}\left[L(\mathbf{W}_{t}) \middle| \mathcal{H}_{t-1}\right] \le L(\mathbf{W}_{t-1}) &- \lrtrandom \left\langle \nabla L(\mathbf{W}_{t-1}), \mathbb{E}\left[\indicatoraceptance_t \widehat{\mathbf{G}}_t \middle| \mathcal{H}_{t-1}\right] \right\rangle \nonumber \\
    &+ \frac{\ell \lrtrandom^2}{2} \mathbb{E}\left[\indicatoraceptance_t \|\widehat{\mathbf{G}}_t\|_2^2 \middle| \mathcal{H}_{t-1}\right].
\end{align}
To simplify \eqref{eq:exp_descent_step_v2}, we analyze each term individually, starting with the expectation in the cross-term $\mathbb{E}\left[\indicatoraceptance_t \widehat{\mathbf{G}}_t \middle| \mathcal{H}_{t-1}\right]$. By applying the law of total expectation, we have
\begin{align}
    \mathbb{E}\left[\indicatoraceptance_t \widehat{\mathbf{G}}_t \middle| \mathcal{H}_{t-1}\right] 
    &= \mathbb{E}\left[\indicatoraceptance_t \widehat{\mathbf{G}}_t \middle| \mathcal{H}_{t-1}, \acceptancet\right] \Pr(\acceptancet \mid \mathcal{H}_{t-1})  + \mathbb{E}\left[\indicatoraceptance_t \widehat{\mathbf{G}}_t \middle| \mathcal{H}_{t-1}, \acceptancet^c\right] \Pr(\acceptancet^c \mid \mathcal{H}_{t-1}) \nonumber \\
    &\overset{(a)}{=} \mathbb{E}\left[\widehat{\mathbf{G}}_t \middle| \mathcal{H}_{t-1}, \acceptancet\right] \Pr(\acceptancet \mid \mathcal{H}_{t-1}) + 0 \nonumber \\
    &\overset{(b)}{=} \pat \nabla L(\mathbf{W}_{t-1}), \label{eq:exp_grad_term_v2}
\end{align}
where $(a)$ follows because the indicator $\indicatoraceptance_t$ is equal to $1$ given $\acceptancet$ and $0$ given $\acceptancet^c$. Step $(b)$ follows from the conditional unbiasedness assumption $\mathbb{E}[\widehat{\mathbf{G}}_t \mid \mathcal{H}_{t-1}, \acceptancet] = \nabla L(\mathbf{W}_{t-1})$ and from the definition $\Pr(\acceptancet \mid \mathcal{H}_{t-1})=\pat$ in \eqref{eq:PA_t_and_eq:MSE_t}.

For the third term in \eqref{eq:exp_descent_step_v2}, we evaluate $\mathbb{E}\left[\indicatoraceptance_t \|\widehat{\mathbf{G}}_t\|_2^2 \middle| \mathcal{H}_{t-1}\right]$. Note that
\begin{align}
    &\mathbb{E}\left[\indicatoraceptance_t \|\widehat{\mathbf{G}}_t\|_2^2 \middle| \mathcal{H}_{t-1}\right] \nonumber \\
    &\quad = \mathbb{E}\left[\indicatoraceptance_t \|\widehat{\mathbf{G}}_t\|_2^2 \middle| \mathcal{H}_{t-1}, \acceptancet\right] \Pr(\acceptancet \mid \mathcal{H}_{t-1}) \nonumber + \mathbb{E}\left[\indicatoraceptance_t \|\widehat{\mathbf{G}}_t\|_2^2 \middle| \mathcal{H}_{t-1}, \acceptancet^c\right] \Pr(\acceptancet^c \mid \mathcal{H}_{t-1}) \nonumber \\
    &\quad \overset{(a)}{=} \mathbb{E}\left[\|\widehat{\mathbf{G}}_t\|_2^2 \middle| \mathcal{H}_{t-1}, \acceptancet\right] \pat + 0 \nonumber \\
    &\quad \overset{(b)}{=} \bigg( \|\mathbb{E}[\widehat{\mathbf{G}}_t \mid \mathcal{H}_{t-1}, \acceptancet]\|_2^2  + \mathbb{E}\left[\|\widehat{\mathbf{G}}_t - \mathbb{E}[\widehat{\mathbf{G}}_t \mid \mathcal{H}_{t-1}, \acceptancet]\|_2^2 \middle| \mathcal{H}_{t-1}, \acceptancet\right] \bigg) \pat \nonumber \\
    &\quad \overset{(c)}{=} \left( \|\nabla L(\mathbf{W}_{t-1})\|_2^2 + \mathbb{E}\left[\|\widehat{\mathbf{G}}_t - \nabla L(\mathbf{W}_{t-1})\|_2^2 \middle| \mathcal{H}_{t-1}, \acceptancet\right] \right) \pat \nonumber \\
    &\quad \overset{(d)}{=} \left( \|\nabla L(\mathbf{W}_{t-1})\|_2^2 + \mset \right) \pat, \label{eq:exp_norm_sq_v2}
\end{align}
where $(a)$ uses the fact that $\indicatoraceptance_t=1$ given $\acceptancet$ and $\indicatoraceptance_t=0$ otherwise, together with the definition $\Pr(\acceptancet \mid \mathcal{H}_{t-1})=\pat$ in \eqref{eq:PA_t_and_eq:MSE_t}. Step $(b)$ applies the decomposition $\mathbb{E}[\|\mathbf{X}\|_2^2] = \|\mathbb{E}[\mathbf{X}]\|_2^2 + \mathbb{E}[\|\mathbf{X} - \mathbb{E}[\mathbf{X}]\|_2^2]$ to the conditional estimator. In $(c)$, we substitute the conditional unbiasedness property $\mathbb{E}[\widehat{\mathbf{G}}_t \mid \mathcal{H}_{t-1}, \acceptancet] = \nabla L(\mathbf{W}_{t-1})$. Finally, $(d)$ follows directly from the conditional definition of the equilibrium mean squared error $\mset$ in \eqref{eq:PA_t_and_eq:MSE_t}.

By substituting \eqref{eq:exp_grad_term_v2} and \eqref{eq:exp_norm_sq_v2} into \eqref{eq:exp_descent_step_v2}, we obtain
\begin{align}
    \mathbb{E}\left[L(\mathbf{W}_{t}) \middle| \mathcal{H}_{t-1}\right] 
    &\le L(\mathbf{W}_{t-1}) - \lrtrandom \pat \|\nabla L(\mathbf{W}_{t-1})\|_2^2  + \frac{\ell \lrtrandom^2 \pat}{2} \left( \|\nabla L(\mathbf{W}_{t-1})\|_2^2 + \mset \right) \nonumber \\
    &= L(\mathbf{W}_{t-1}) - \lrtrandom \pat \left( 1 - \frac{\ell \lrtrandom}{2} \right) \|\nabla L(\mathbf{W}_{t-1})\|_2^2  + \frac{\ell \lrtrandom^2}{2} \pat \mset. \label{eq:final_expected_descent_v2}
\end{align}

To further analyze the expected descend of the loss function, we need to study two disjoint cases, wherein the acceptance parameter $\eta_t$ is determined differently. More precisely, we partition the rounds $t \in \{0, \dots, T\}$ into two disjoint sets, the adaptive regime ${\mathcal{T}_a = \{ t \mid \eta_t > \eta_{\min} \}}$ and the saturated regime ${\mathcal{T}_s = \{ t \mid \eta_t = \eta_{\min} \}}$, and for each of them we rearrange the bound in~\eqref{eq:final_expected_descent_v2}.

\textbf{Case 1 ($t \in \mathcal{T}_a$):} Recall that in the idealized execution analyzed in Theorem~\ref{thm:convergence}, the DC computes the target variance as 
\begin{align}\label{eq:target_variance}
    \targetvariance
    =
    \min\{\sigmax,\max\{\sigmin,\, c\|\nabla L(\mathbf{W}_{t-1})\|_2^2\}\},
\end{align}
and selects the threshold $\eta_t \in [\eta_{\min}, \eta_{\max}]$ as the smallest value such that
\begin{align}\label{eq:mse_target_matching}
    \mset = \targetvariance.
\end{align}
Note that based on this threshold selection rule, for $t \in \mathcal{T}_a$, we have $\eta_t>\eta_{\min}$, and hence
\begin{align}\label{eq:mset_adaptive_regime}
    \mset = \targetvariance \leq c \|\nabla L(\mathbf{W}_{t-1})\|_2^2.
\end{align}
This is due to the fact that otherwise, meaning if $\targetvariance = \sigmin$, we would have $\eta_t = \eta_{\min}$ based on the threshold selection process, which contradicts our assumption of $t \in \mathcal{T}_a$.

Also, note that we select the initial learning rate $b_0$ such that $b_0 \le 1/(\ell(1+c))$. Since $\lrtrandom \le b_0$ for all $t$, this choice guarantees
\begin{equation} \label{eq:alpha_conditions}
    1 - \frac{\ell \lrtrandom(1+c)}{2} \ge \frac{1}{2}.
\end{equation}

Now, based on \eqref{eq:final_expected_descent_v2}, we have
\begin{align} 
    \mathbb{E}\left[L(\mathbf{W}_{t}) \middle| \mathcal{H}_{t-1}\right]  
    &\overset{(a)}{\leq} L(\mathbf{W}_{t-1}) - \lrtrandom \pat \left( 1 - \frac{\ell \lrtrandom}{2} \right) \|\nabla L(\mathbf{W}_{t-1})\|_2^2  + \frac{\ell \lrtrandom^2}{2} \pat c \|\nabla L(\mathbf{W}_{t-1})\|_2^2 \nonumber\\
    &= L(\mathbf{W}_{t-1}) - \lrtrandom \pat \left( 1 - \frac{\ell \lrtrandom (1+c)}{2} \right) \|\nabla L(\mathbf{W}_{t-1})\|_2^2 \nonumber \\
    &\overset{(b)}{\leq} L(\mathbf{W}_{t-1}) - \frac{\lrtrandom \pmin}{2} \|\nabla L(\mathbf{W}_{t-1})\|_2^2, \label{eq:descent_adaptive}
\end{align}
where $(a)$ follows from substituting the upper bound of the estimation error $\mset$ from~\eqref{eq:mset_adaptive_regime} into~\eqref{eq:final_expected_descent_v2}. Step $(b)$ follows from applying the learning rate condition in \eqref{eq:alpha_conditions} and utilizing the fact that the acceptance probability $\pat$ is bounded from below by a minimum probability $\pmin$, ensuring ${\pat \ge \pmin}$.

\textbf{Case 2 ($t \in \mathcal{T}_s$):} In this case, by definition we have $\eta_t = \eta_{\min}$, and thus we have 
\begin{align}\label{eq:mset_saturated_regime}
    \mset = \sigmin.
\end{align}
Based on \eqref{eq:final_expected_descent_v2}, we arrive at
\begin{align} 
    \mathbb{E}\left[L(\mathbf{W}_{t}) \middle| \mathcal{H}_{t-1}\right]  
    &\overset{(a)}{\leq} L(\mathbf{W}_{t-1}) - \lrtrandom \pat \left( 1 - \frac{\ell \lrtrandom}{2} \right) \|\nabla L(\mathbf{W}_{t-1})\|_2^2 + \frac{\ell \lrtrandom^2}{2} \pat \sigmin \nonumber \\
    &\overset{(b)}{\leq} L(\mathbf{W}_{t-1}) - \frac{\lrtrandom \pmin}{2} \|\nabla L(\mathbf{W}_{t-1})\|_2^2 + \frac{\ell \sigmin}{2} \pat \lrtrandom^2, \label{eq:descent_saturated}
\end{align}
where $(a)$ follows from \eqref{eq:mset_saturated_regime}. Step $(b)$ follows from the fact that, based on \eqref{eq:alpha_conditions}, we have
\begin{align}\label{eq:learning_rate_bound_saturated}
    1 - \frac{\ell \lrtrandom}{2} \geq 1 - \frac{\ell \lrtrandom(1+c)}{2} \ge \frac{1}{2},
\end{align}
and the acceptance probability $\pat$ is bounded from below by a minimum probability $\pmin$, ensuring $\pat \ge \pmin$.

\textbf{Unified Bound:}  Using the indicator function $\indicatorsaturate_t$ to distinguish the event that round $t$ is accepted in the saturated regime, we can aggregate \eqref{eq:descent_adaptive} and~\eqref{eq:descent_saturated} into a single inequality for all $t$, and obtain
\begin{align} 
    \mathbb{E}\left[L(\mathbf{W}_{t}) \middle| \mathcal{H}_{t-1}\right] 
    &\le L(\mathbf{W}_{t-1}) - \frac{\lrtrandom \pmin}{2} \|\nabla L(\mathbf{W}_{t-1})\|_2^2 + \frac{\ell \sigmin}{2} \lrtrandom^2 \mathbb{E}\left[\indicatorsaturate_t \middle| \mathcal{H}_{t-1}\right]. \label{eq:bounded_descent_v2}
\end{align}
Indeed, if $t\in\mathcal{T}_a$, then $\eta_t>\eta_{\min}$ and $\mathbb{E}[\indicatorsaturate_t \mid \mathcal{H}_{t-1}]=0$. If $t\in\mathcal{T}_s$, then $\eta_t=\eta_{\min}$ and $\mathbb{E}[\indicatorsaturate_t \mid \mathcal{H}_{t-1}]=\Pr(\acceptancet\mid\mathcal{H}_{t-1})=\pat$, which matches the last term in \eqref{eq:descent_saturated}. 

Equation~\eqref{eq:bounded_descent_v2} provides a unified descent step; in the adaptive regime, the error term vanishes due to the dynamic variance control, while in the saturated regime, the descent is subject to the constant variance $\sigmin$.
We rearrange \eqref{eq:bounded_descent_v2} to get
\begin{equation} \label{eq:rearranged_descent_v2}
    \frac{\lrtrandom \pmin}{2} \|\nabla L(\mathbf{W}_{t-1})\|_2^2 \le L(\mathbf{W}_{t-1}) - \mathbb{E}\left[L(\mathbf{W}_{t}) \middle| \mathcal{H}_{t-1}\right] + \frac{\ell \sigmin}{2} \lrtrandom^2 \mathbb{E}\left[\indicatorsaturate_t \middle| \mathcal{H}_{t-1}\right].
\end{equation}

Next, we take the expectation of both sides of \eqref{eq:rearranged_descent_v2} with respect to all the randomness in $\mathcal{H}_{t-1}$. Since $\lrtrandom$ is $\mathcal{H}_{t-1}$-measurable, this yields
\begin{align}\label{eq:total_expectation_descent}
    \frac{\pmin}{2} \mathbb{E}\left[\lrtrandom \|\nabla L(\mathbf{W}_{t-1})\|_2^2\right] 
    &\hspace{-1pt}\le\hspace{-1pt} \mathbb{E}[L(\mathbf{W}_{t-1})] \hspace{-1pt}-\hspace{-1pt} \mathbb{E}_{\mathcal{H}_{t-1}}\hspace{-2pt}\left[\mathbb{E}\left[L(\mathbf{W}_{t}) \middle| \mathcal{H}_{t-1}\right]\right]  \hspace{-1pt}+ \hspace{-1pt}\frac{\ell \sigmin}{2} \mathbb{E}\left[ \lrtrandom^2 \mathbb{E}\left[\indicatorsaturate_t \middle| \mathcal{H}_{t-1}\right] \right] \nonumber \\
    &\overset{(a)}{=} \mathbb{E}[L(\mathbf{W}_{t-1})] - \mathbb{E}[L(\mathbf{W}_{t})] + \frac{\ell \sigmin}{2} \mathbb{E}\left[ \indicatorsaturate_t \lrtrandom^2 \right],
\end{align}
where $(a)$ follows from the law of total expectation and the fact that $\lrtrandom$ is $\mathcal{H}_{t-1}$-measurable.

Summing up \eqref{eq:total_expectation_descent} for $t=0$ to $T$, we obtain
\begin{equation} \label{eq:summed_descent_v2}
    \frac{\pmin}{2} \sum_{t=0}^{T} \mathbb{E}\left[\lrtrandom \|\nabla L(\mathbf{W}_{t-1})\|_2^2\right] \le \sum_{t=0}^{T} \left( \mathbb{E}[L(\mathbf{W}_{t-1})] - \mathbb{E}[L(\mathbf{W}_{t})] \right) + \frac{\ell \sigmin}{2} \sum_{t=0}^{T} \mathbb{E}\left[ \indicatorsaturate_t \lrtrandom^2 \right].
\end{equation}
Next, we evaluate the two sums on the right side of \eqref{eq:summed_descent_v2}. Note that the initial state $\mathbf{W}_{-1}$ is deterministic, and hence, $\mathbb{E}[L(\mathbf{W}_{-1})] = L(\mathbf{W}_{-1})$. Using the assumption that $L$ is bounded from below by $L^*$, we have
\begin{equation} \label{eq:telescoping_sum_v2}
    \sum_{t=0}^{T} \left( \mathbb{E}[L(\mathbf{W}_{t-1})] - \mathbb{E}[L(\mathbf{W}_{t})] \right) = L(\mathbf{W}_{-1}) - \mathbb{E}[L(\mathbf{W}_{T})] \le L(\mathbf{W}_{-1}) - L^*.
\end{equation}
For the second sum $\frac{\ell \sigmin}{2} \sum_{t=0}^{T} \mathbb{E}\left[ \indicatorsaturate_t \lrtrandom^2 \right]$, we use the update rule of the decay counter described in \algoname\ (Algorithm \ref{alg:avm_dgd}). Recall that $\indicatorsaturate_t$ is an indicator where $\indicatorsaturate_t = 1$ if round $t$ was accepted ($\indicatoraceptance_t = 1$) and the system was in the saturated regime ($\eta_t = \eta_{\min}$), and $0$ otherwise. Furthermore, recall that the decay counter at round $t$ is defined as $\tau_t = \sum_{k=0}^{t} \indicatorsaturate_k$, and the learning rate is determined by $\lrtrandom = \frac{b_0}{\sqrt{\tau_{t-1} + 1}}$, with the initial value of $\tau_{-1} = 0$.

Due to these definitions, the counter increments and the learning rate decays in iteration $t$ only when $\indicatorsaturate_{t-1} = 1$, and otherwise we have $\tau_{t-1}=\tau_{t-2}$ and hence $B_t=B_{t-1}$. Let $S =\tau_T= \sum_{t=0}^{T} \indicatorsaturate_t$ denote the total number of saturated rounds up to iteration $T$, and note that $S \le T+1$. While $B_t$ is random, and may decay (when $\indicatorsaturate_{t-1} = 1$) or remain constant (when $\indicatorsaturate_{t-1} = 0$), the product $B_t^2 \indicatorsaturate_t$ takes each value of $\left(\frac{b_0}{\sqrt{k+1}}\right)^2$ for $k=0,\dots, S-1$, exactly once. Therefore, the only remaining randomness in the second summation is due to $S$, which can deterministically upper-bounded by $S\leq T+1$. 

Therefore, the sum of the squared learning rates over these rounds exactly forms a harmonic series, which is bounded by 
\begin{align} \label{eq:alpha_sq_sum_v2}
    \sum_{t=0}^{T} \indicatorsaturate_t \lrtrandom^2 &= b_0^2 \sum_{k=1}^{S} \frac{1}{k}  = b_0^2 \left( 1 + \sum_{k=2}^{S} \frac{1}{k} \right) \nonumber \\
    &\overset{(a)}{\le} b_0^2 \left( 1 + \sum_{k=2}^{S}\int_{k-1}^{k} \frac{1}{x} dx \right)\nonumber\\
    &= b_0^2 \left( 1 + \int_{1}^{S} \frac{1}{x} dx \right)\nonumber \\
    &= b_0^2 (1 + \ln S) \le b_0^2 (1 + \ln(T+1)),
\end{align}
where $(a)$ follows because $f(x) = 1/x$ is monotonically decreasing for $x>0$, ensuring that for every integer $k \ge 2$ we have
\begin{equation} \label{eq:integral_comparison_harmonic}
    \frac{1}{k} \le \int_{k-1}^{k} \frac{1}{x} dx.
\end{equation}
Substituting \eqref{eq:telescoping_sum_v2} and the expectation of the bounded sum from \eqref{eq:alpha_sq_sum_v2} into \eqref{eq:summed_descent_v2} yields
\begin{equation} \label{eq:accumulated_bound_v2}
    \frac{\pmin}{2} \sum_{t=0}^{T} \mathbb{E}\left[\lrtrandom \|\nabla L(\mathbf{W}_{t-1})\|_2^2\right] \le (L(\mathbf{W}_{-1}) - L^*) + \frac{\ell \sigmin b_0^2}{2} (1 + \ln(T+1)).
\end{equation}

To establish the final convergence rate, we use the property that the minimum value of a nonnegative sequence is always bounded above by any deterministic weighted average of that sequence. According to Algorithm~\ref{alg:avm_dgd}, the decay in the learning rate only occurs when an update is accepted in the saturated regime; otherwise, the learning rate does not change. Since the decay counter increments at most once per round, we have the lower bound $\lrtrandom \ge b_0 / \sqrt{t+1}$ for all $t \in \{0, \dots, T\}$. Therefore, since $\|\nabla L(\mathbf{W}_{t-1})\|_2^2$ is nonnegative, we have
\begin{align} \label{eq:weighted_average_trick_v2}
    \sum_{t=0}^{T} \mathbb{E}\left[\lrtrandom \|\nabla L(\mathbf{W}_{t-1})\|_2^2\right]
    &\ge
    b_0\sum_{t=0}^{T}\frac{1}{\sqrt{t+1}}
    \mathbb{E}\left[\|\nabla L(\mathbf{W}_{t-1})\|_2^2\right] \nonumber \\
    &\ge
    \left(
    \min_{0 \le t \le T}
    \mathbb{E}\left[\|\nabla L(\mathbf{W}_{t-1})\|_2^2\right]
    \right)
    b_0\sum_{t=0}^{T}\frac{1}{\sqrt{t+1}} .
\end{align}
Next, we lower bound the deterministic weight sum. Note that
\begin{align} \label{eq:alpha_sum_integral_v2}
    b_0\sum_{t=0}^{T}\frac{1}{\sqrt{t+1}}
    &\overset{(a)}{\ge}
    b_0 \sum_{t=0}^{T} \int_{t}^{t+1} \frac{1}{\sqrt{x+1}}\,dx \nonumber \\
    &= b_0 \int_{0}^{T+1}\frac{1}{\sqrt{x+1}}\,dx \nonumber \\
    &= 2b_0(\sqrt{T+2}-1),
\end{align}
where $(a)$ follows because $f(x)=1/\sqrt{x+1}$ is monotonically decreasing on $[0,\infty)$, ensuring that for every integer $t \ge 0$ we have
\begin{equation} \label{eq:integral_comparison_v2}
    \frac{1}{\sqrt{t+1}} \ge \int_t^{t+1}\frac{1}{\sqrt{x+1}}\,dx.
\end{equation}
Combining \eqref{eq:weighted_average_trick_v2} and \eqref{eq:alpha_sum_integral_v2}, we obtain
\begin{equation} \label{eq:intermediate_convergence_bound}
    \min_{0 \le t \le T}
    \mathbb{E}\left[\|\nabla L(\mathbf{W}_{t-1})\|_2^2\right]
    \le
    \frac{
    \sum_{t=0}^{T}
    \mathbb{E}\left[\lrtrandom \|\nabla L(\mathbf{W}_{t-1})\|_2^2\right]
    }{
    2 b_0 (\sqrt{T+2}-1)
    }.
\end{equation}

Therefore, substituting the upper bound of the sum from \eqref{eq:accumulated_bound_v2} into \eqref{eq:intermediate_convergence_bound}, we obtain the final convergence rate
\begin{equation} \label{eq:final_convergence_rate_v2}
    \min_{0 \le t \le T}
    \mathbb{E}\left[\|\nabla L(\mathbf{W}_{t-1})\|_2^2\right]
    \le
    \frac{
    2(L(\mathbf{W}_{-1}) - L^*) + \ell \sigmin b_0^2 (1 + \ln(T+1))
    }{
    2 \pmin b_0 (\sqrt{T+2}-1)
    }.
\end{equation}

As $T \to \infty$, the denominator $2\pmin b_0(\sqrt{T+2}-1)$ grows as $\Theta(\sqrt{T})$ and the numerator grows as $\Theta(\ln T)$. Factoring out the leading terms and constants from \eqref{eq:final_convergence_rate_v2}, we obtain the final asymptotic upper bound
\begin{equation} \label{eq:final_rate_bound_v2}
    \min_{0 \le t \le T} \mathbb{E}\left[\|\nabla L(\mathbf{W}_{t-1})\|_2^2\right] \le \mathcal{O}\left( \frac{\ln T}{\sqrt{T}}  \right) .
\end{equation}

\section{Experimental Details}
\label{app:exp_details}

In this section, we provide additional details for the experiments in Section~\ref{Sec:Experiments}. We also include a one-dimensional experiment with $\Nnodes=10$ agents to illustrate a setting with more than two participating nodes.

\begin{figure*}[t]
    \centering
    \begin{subfigure}[t]{0.48\textwidth}
        \centering
        \includegraphics[width=\linewidth]{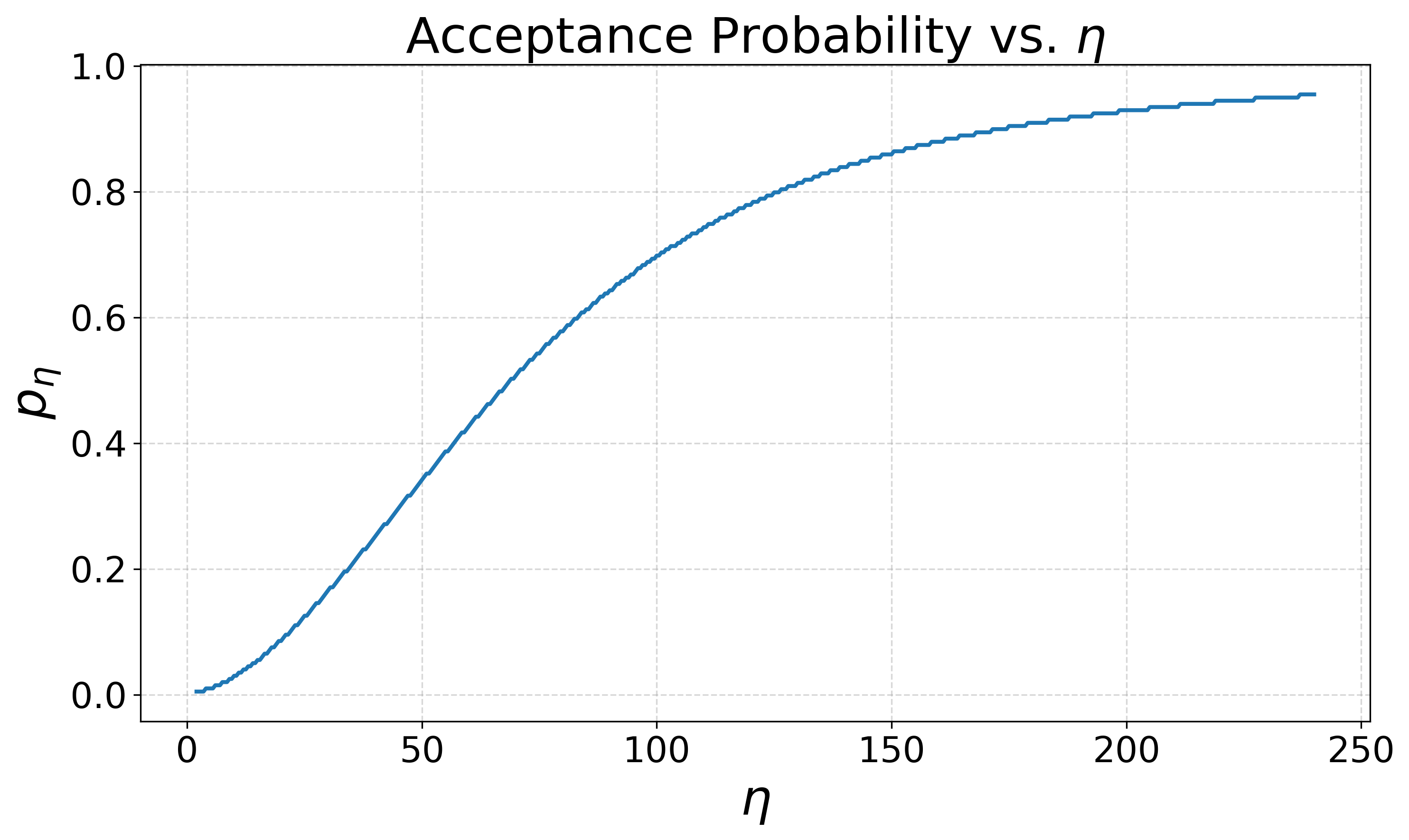}
        \caption{}
        \label{fig:eq_3d_pa}
    \end{subfigure}
    \hfill
    \begin{subfigure}[t]{0.48\textwidth}
        \centering
        \includegraphics[width=\linewidth]{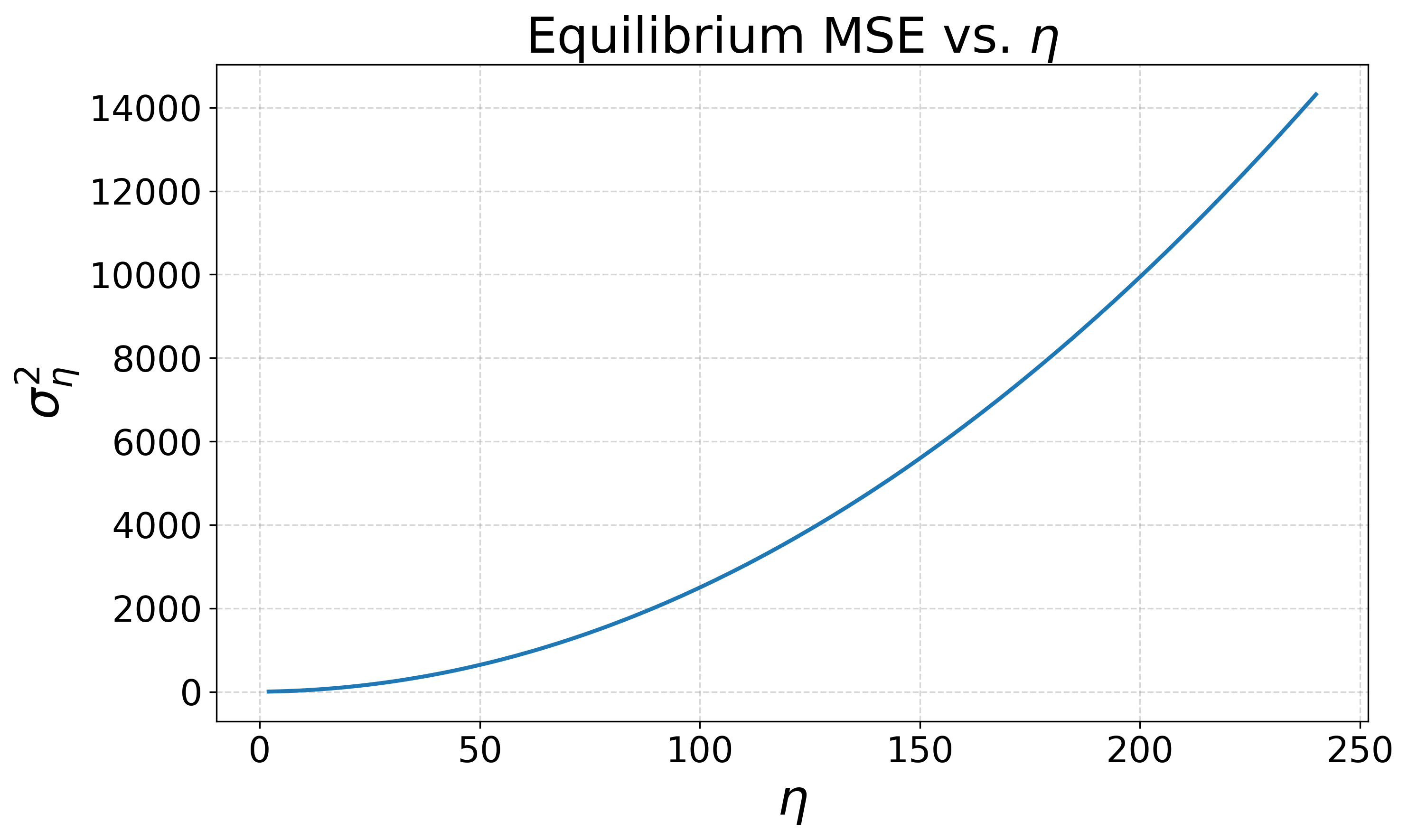}
        \caption{}
        \label{fig:eq_3d_mse}
    \end{subfigure}
    \caption{Equilibrium characterization for the three-dimensional experiment in Section \ref{Sec:Experiments}. The adversarial utility is $\mathsf{U}_{\mathrm{AD}}=\log(\mse)+0.03\log(\pa)$. Using the results of~\cite{nodehi2026gamevector}, we characterize the equilibrium values of $p_{\eta}$ and $\sigma_{\eta}^2$ for each $\eta \in [\eta_{\min},\eta_{\max}]$, where $\eta_{\min}=2$ and $\eta_{\max}=240$. Subfigure~(a) shows the equilibrium acceptance probability $p_{\eta}$, and subfigure~(b) shows the corresponding equilibrium mean squared error $\sigma_{\eta}^2$.}
    \label{fig:eq_3d_characterization}
\end{figure*}

Throughout the experiments, we use prior game-of-coding results to obtain the equilibrium acceptance
probability $p_{\eta}$, the equilibrium mean squared error $\sigma^2_{\eta}$, the estimator
$\mathrm{est}(\cdot)$, and the adversarial noise strategy for each value of $\eta$. These quantities are
not derived in this paper; rather, they are taken from the corresponding game-of-coding
characterizations in the prior literature~\cite{GoCJournal, nodehi2026gamevector, GoDSybil, nodehi2025unknown, nodehi2026game}. For instance, for $d=3$, the evolution of $p_{\eta}$ and $\sigma_{\eta}^2$ as functions of $\eta$ are reported in Fig.~\ref{fig:eq_3d_characterization}. These functions are derived based on the work of~\cite{nodehi2026game}  for the specific utility function $\mathsf{U}_{\mathrm{AD}}=\log(\mse)+0.03\log(\pa)$, which is used for the three-dimensional experiment reported in Section~\ref{Sec:Experiments}. Note that in \cite{nodehi2026game}, it has been proved that the behavior of $p_{\eta}$ and $\sigma_{\eta}^2$ depend on the dimension and adversary's utility function, but not the objective function to be optimized. Nevertheless, characterizing the equilibrium noise strategy and the
associated estimator is a separate game-theoretic problem and is beyond the scope of this paper.

For scalar computations $(d=1)$, the general multi-node case has been characterized in prior
game-of-coding works~\cite{GoCJournal,GoDSybil}. For vector-valued computations $(d>1)$,
the two-node case, consisting of one honest node and one adversarial node, has been characterized
in~\cite{nodehi2026gamevector}. This is the vector-valued setting used in our experiments in
Section~4. The general vector-valued case with more than two nodes remains open; once the
corresponding equilibrium curve and estimator are available, VISTA can be applied to that setting
without any changes.

In particular, for each value of $\eta$, we use the equilibrium noise strategy and the corresponding
equilibrium estimator induced by the cited game-of-coding characterization, and then use this
estimator as $\mathrm{est}(\cdot)$ inside VISTA. Thus, the contribution of this paper is not to solve the
underlying single-round game again, but to show how the resulting equilibrium quantities can be
used to design an adaptive threshold and learning-rate schedule for multi-round optimization. As
mentioned earlier, we consider the complete-information setting in which this equilibrium curve is
available. When the adversary model or utility is unknown, the same direction can be combined with
the learning-based techniques in~\cite{nodehi2025unknown,nodehi2026game}, which we leave for
future work.

\begin{figure}
    \centering
    \includegraphics[width=0.75\linewidth]{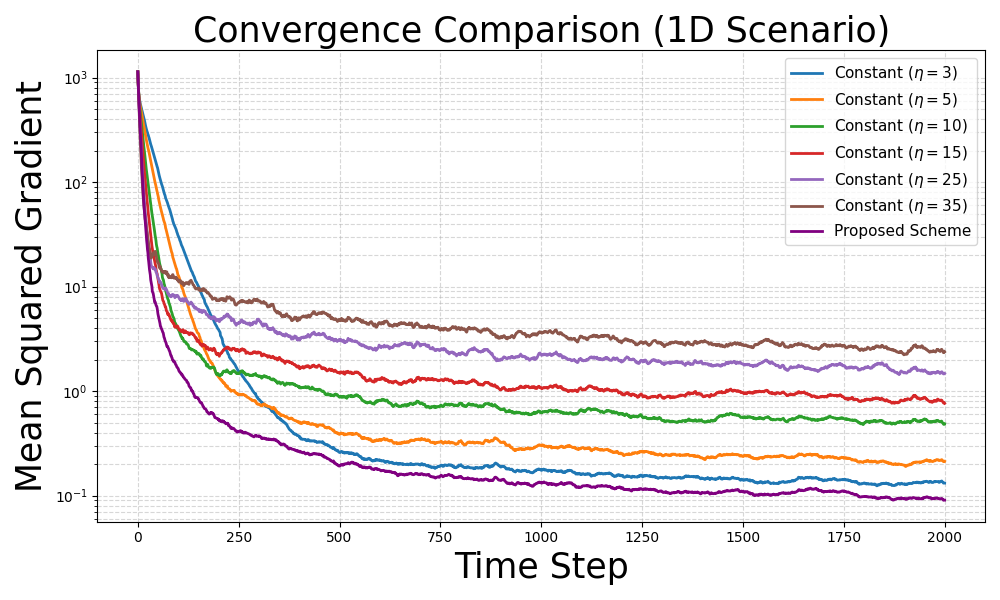}
    \caption{\small{Convergence in the one-dimensional case. Larger values of $\eta$ improve early progress but worsen final accuracy, while the proposed scheme achieves the best performance.}}
    \label{fig:experimental_results_1d}
\end{figure}

\begin{figure*}[htbp]
    \centering
    \begin{subfigure}[t]{0.48\textwidth}
        \centering
        \includegraphics[width=\linewidth]{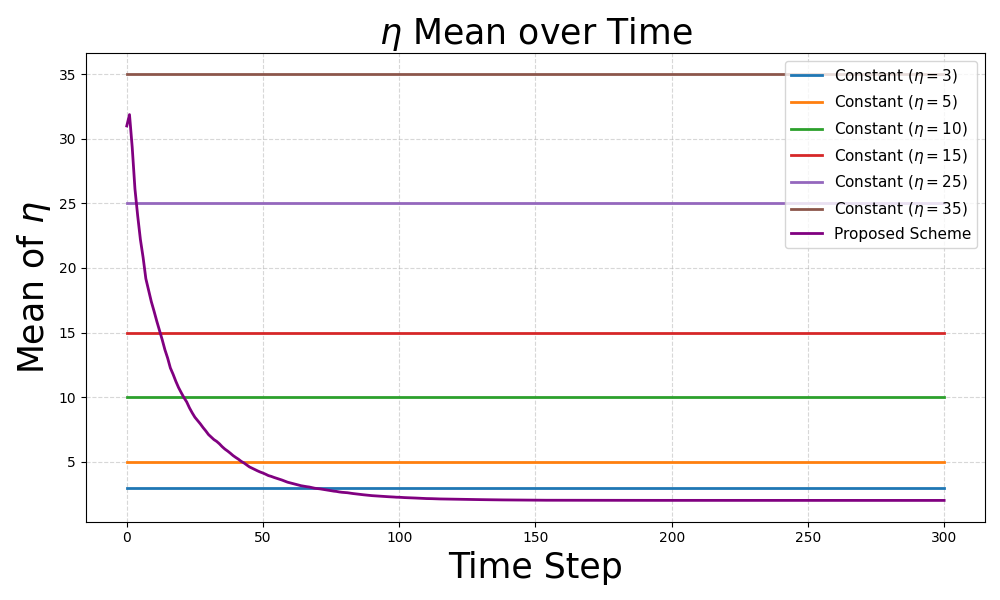}
        \caption{}
        \label{fig:eta_results_1d}
    \end{subfigure}
    \hfill
    \begin{subfigure}[t]{0.48\textwidth}
        \centering
        \includegraphics[width=\linewidth]{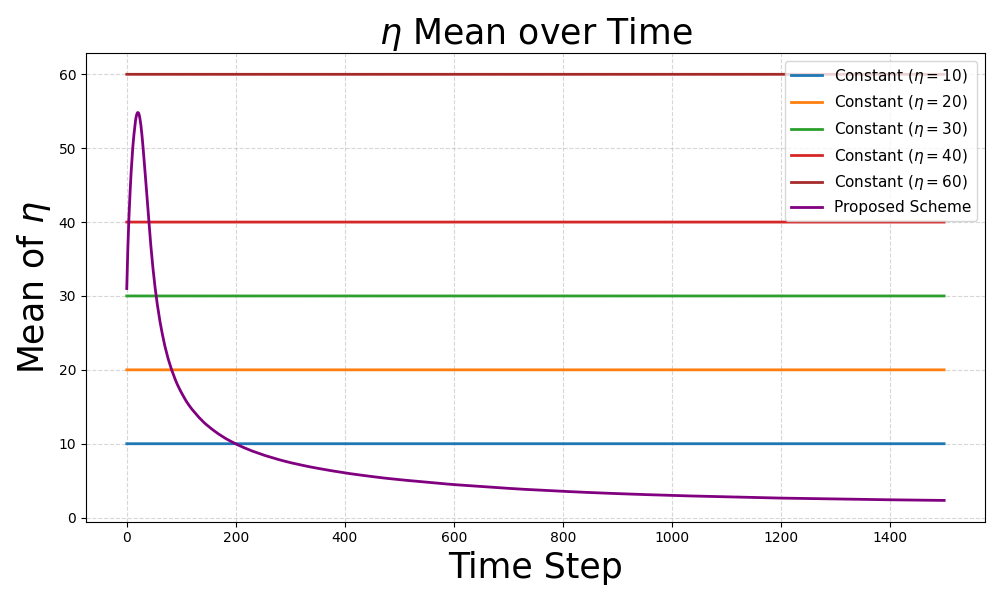}
        \caption{}
        \label{fig:eta_results_3d}
    \end{subfigure}
    \caption{\small{Dynamic evolution of the acceptance threshold $\eta_t$ across the experiments. (a) One-dimensional experiment. (b) Three-dimensional experiment. In both cases, the proposed scheme adapts $\eta_t$ over time, starting from a larger value and gradually tightening the acceptance rule.}}
    \label{fig:combined_eta_results_1}
\end{figure*}

Consider a one-dimensional example, where $L(w)=10w\sin(w/10)$, $\Nnodes=10$ agents, and up to $9$ adversarial nodes. The adversarial utility is $\mathsf{U}_{\mathrm{AD}}=\log(\mse)+0.1\log(\pa)$, and we set $(c,b_0)=(1,0.1)$. We repeat the experiment over $500$ independent runs and report the average performance. Fig.~\ref{fig:experimental_results_1d} shows a clear \textit{tortoise and hare effect}. Smaller constant values of $\eta$ are more conservative and therefore slower initially, while larger values of $\eta$ descend faster at the beginning but eventually suffer from a higher error floor due to larger adversarial distortion. The proposed scheme, shown by the purple curve, combines the benefits of both behaviors by descending quickly at the beginning and achieving the best final performance. The corresponding evolution of $\eta_t$ is shown in Fig.~\ref{fig:combined_eta_results_1}\subref{fig:eta_results_1d}.

Table~\ref{tab:exp_details} summarizes the main implementation details used in the experiments in Section~\ref{Sec:Experiments}, together with the one-dimensional experiment reported in this appendix. For all experiments, we set ${\Delta=1}$, $\eta_{\min}=2$, and, similar to Adam~\cite{kingma2015adam}, we set $\beta=0.9$. The initial threshold is chosen as ${\eta_0=(\eta_{\min}+\eta_{\max})/2}$. The value of $\eta_{\max}$ is selected separately for each experiment so that the acceptance-probability range of $(0,1]$ is properly covered.

In all experiments, we instantiate the honest noise by drawing it uniformly from the $\ndim$-dimensional Euclidean ball of radius $\Delta$. For the adversarial noise, we use the corresponding game-of-coding equilibrium construction. In the one-dimensional experiment, we use the scalar game-of-coding procedures~\cite{GoCJournal,GoDSybil}, where the equilibrium and adversarial noise strategy are characterized through their Algorithm~1 and Algorithm~2. For the three-dimensional and deep-learning experiments, we use the vector-valued game-of-coding procedure~\cite{nodehi2026gamevector}, where Algorithm~1 characterizes the equilibrium for each value of $\eta$ and Algorithm~2 constructs the corresponding adversarial noise strategy.

The adversarial utilities in Table~\ref{tab:exp_details} follow the same logarithmic form studied in prior game-of-coding works. The coefficient multiplying $\log(\pa)$ controls the relative importance of acceptance probability compared with estimation error. Since the scale of $\mse$ and the sensitivity of $\pa$ depend on the dimension, objective, and noise geometry, this coefficient is chosen so that the induced game exhibits a meaningful tradeoff between acceptance probability and estimation error over the selected range of~$\eta$. It is worth emphasizing that the proposed algorithm $\algoname$, is not restricted to this logarithmic utility form.

For the deterministic synthetic experiments, we initialize the one-dimensional experiment at ${w_{-1}=40}$ and the three-dimensional experiment at $\mathbf{w}_{-1}=(10,20,30)$. For the neural-network experiments, we initialize the model parameters using Kaiming initialization~\cite{he2015delving}. The batch size is $128$ for both LeNet on MNIST and ResNet-18 on CIFAR-10.

\begin{table*}[h]
\centering
\caption{Implementation details for the experiments in Section~\ref{Sec:Experiments}. In all experiments, $\Delta=1$, $\eta_{\min}=2$, $\eta_0=(\eta_{\min}+\eta_{\max})/2$, and $\beta=0.9$.}
\label{tab:exp_details}
\resizebox{\textwidth}{!}{
\begin{tabular}{lcccccccccc}
\toprule
\textbf{Experiment} &
\textbf{$\ndim$} &
\textbf{$T$} &
\textbf{\# runs} &
\textbf{$\eta_{\max}$} &
\textbf{$\eta_0$} &
\textbf{$b_0$} &
\textbf{$c$} &
\textbf{Utility} &
\textbf{Batch size} &
\textbf{Initialization} \\
\midrule
1D synthetic &
$1$ &
$2000$ &
$500$ &
$60$ &
$31$ &
$0.1$ &
$1$ &
$\log(\mse)+0.1\log(\pa)$ &
N/A &
$w_{-1}=40$ \\

3D synthetic &
$3$ &
$10000$ &
$12000$ &
$240$ &
$121$ &
$0.1$ &
$1$ &
$\log(\mse)+0.03\log(\pa)$ &
N/A &
$\mathbf{w}_{-1}=(10,20,30)$ \\

LeNet on MNIST &
$23942$ &
$5000$ &
$16$ &
$24$ &
$13$ &
$0.25$ &
$15$ &
$\log(\mse)+0.0003\log(\pa)$ &
$128$ &
Kaiming~\cite{he2015delving} \\

ResNet-18 on CIFAR-10 &
$11173962$ &
$5000$ &
$16$ &
$80$ &
$41$ &
$0.01$ &
$5$ &
$\log(\mse)+0.00001\log(\pa)$ &
$128$ &
Kaiming~\cite{he2015delving} \\
\bottomrule
\end{tabular}
}
\end{table*}

\begin{figure*}[htbp]
    \centering
    \begin{subfigure}[t]{0.48\textwidth}
        \centering
        \includegraphics[width=\linewidth]{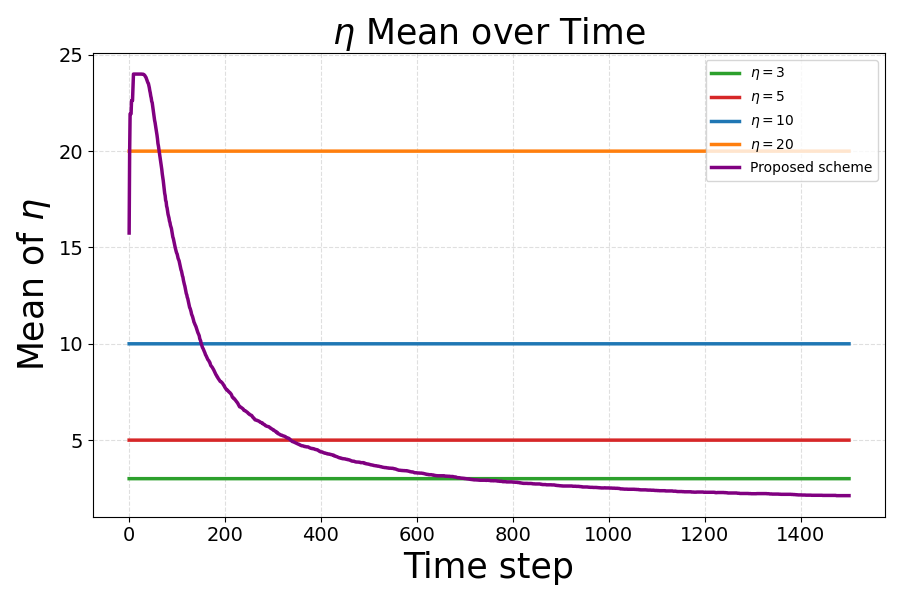}
        \caption{}
        \label{fig:eta_results_lenet}
    \end{subfigure}
    \hfill
    \begin{subfigure}[t]{0.48\textwidth}
        \centering
        \includegraphics[width=\linewidth]{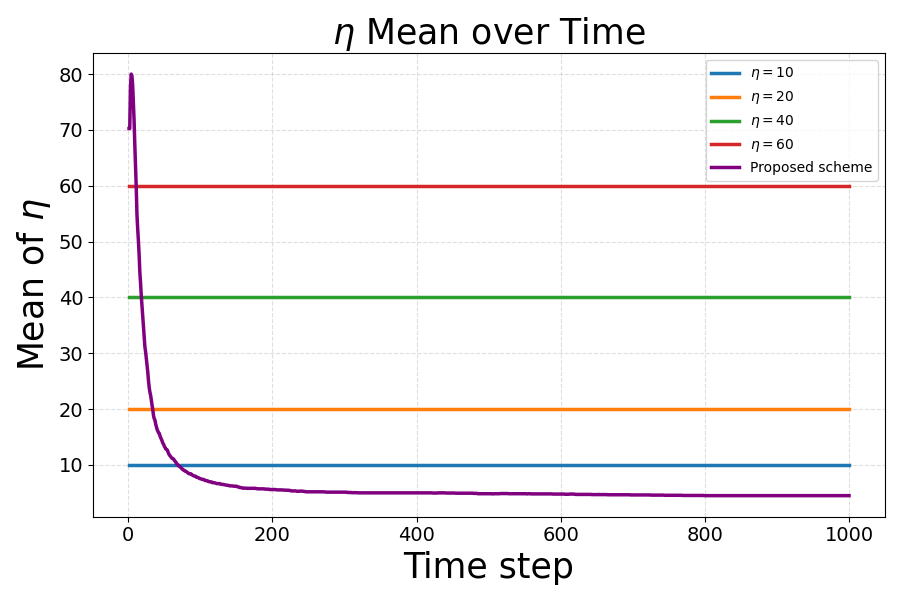}
        \caption{}
        \label{fig:eta_results_resnet}
    \end{subfigure}
    \caption{\small{Dynamic evolution of the acceptance threshold $\eta_t$ across the experiments. (a) LeNet on MNIST. (b) ResNet-18 on CIFAR-10. In both cases, the proposed scheme adapts $\eta_t$ over time, starting from a larger value and gradually tightening the acceptance rule.}}
    \label{fig:combined_eta_results_2}
\end{figure*}

For each experiment, we run \algoname\ for $T$ rounds and repeat the experiment over the number of independent runs reported in Table~\ref{tab:exp_details}. The performance curves for the one-dimensional, three-dimensional, LeNet, and ResNet-18 experiments are shown in Figs.~\ref{fig:experimental_results_1d}, \ref{fig:experimental_results}, \ref{fig:mnist_loss_results}, and~\ref{fig:resnet_loss_results}, respectively. Whenever multiple independent runs are used, the reported curve is the average over those runs. In  figures with shaded regions, the shaded area represents standard deviation across independent runs. The corresponding evolution of the adaptive threshold $\eta_t$ is reported in Fig.~\ref{fig:combined_eta_results_1} and \ref{fig:combined_eta_results_2}.

All experiments are implemented in PyTorch~\citep{paszke2019pytorch} and conducted on a single machine with an NVIDIA RTX 5090 GPU. The synthetic experiments, including all fixed-threshold baselines for each setting, take approximately $30$ minutes per experimental setting. The LeNet and ResNet-18 experiments are run for $T=5000$ rounds over $16$ independent runs, and each of these neural-network experiments takes approximately $6$ hours on the same machine.

\section{Sensitivity to the Hyperparameter \texorpdfstring{$c$}{c}}
\label{app:c_sensitivity}

In this section, we study the role of the hyperparameter $c$ in \algoname. Recall that in Line~\ref{alg:eta_update_line} of Algorithm~\ref{alg:avm_dgd}, the threshold for the next round is selected through 
\begin{equation}
    \eta_{t+1} = \inf\{\eta\in[\eta_{\min},\eta_{\max}]:
        \mathsf{MSE}(g^*_\eta,\eta) = \min\{\sigmax,\max\{\sigmin,\, c\|\widetilde{\mathbf{G}}_{t}\|_2^2\}\}\}.
\end{equation}
Thus, $c$ controls how much estimation error the DC is willing to tolerate relative to the current gradient-signal proxy. A smaller value of $c$ imposes a stricter variance target, which leads to a smaller acceptance threshold and more accurate accepted updates, but it can also push the algorithm more quickly toward the saturated regime $\eta_t=\eta_{\min}$, where updates are accepted less frequently. In contrast, a larger value of $c$ permits a larger variance target, which allows a larger threshold, increases the probability of acceptance, and keeps the algorithm in the adaptive regime for longer. Hence, $c$ acts as a signal-to-noise tolerance parameter: smaller values are more conservative, while larger values are more aggressive.

\begin{figure*}[t]
    \centering
    \begin{subfigure}[t]{0.48\textwidth}
        \centering
        \includegraphics[height=42mm]{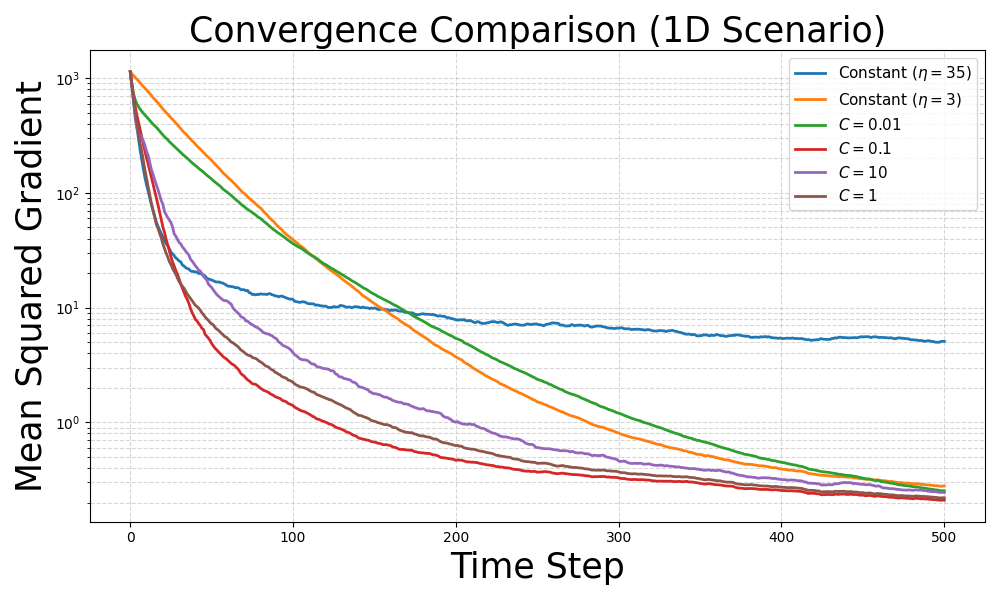}
        \caption{\small{One-dimensional synthetic.}}
        \label{fig:c_sensitivity_perf_1d}
    \end{subfigure}
    \hfill
    \begin{subfigure}[t]{0.48\textwidth}
        \centering
        \includegraphics[height=42mm]{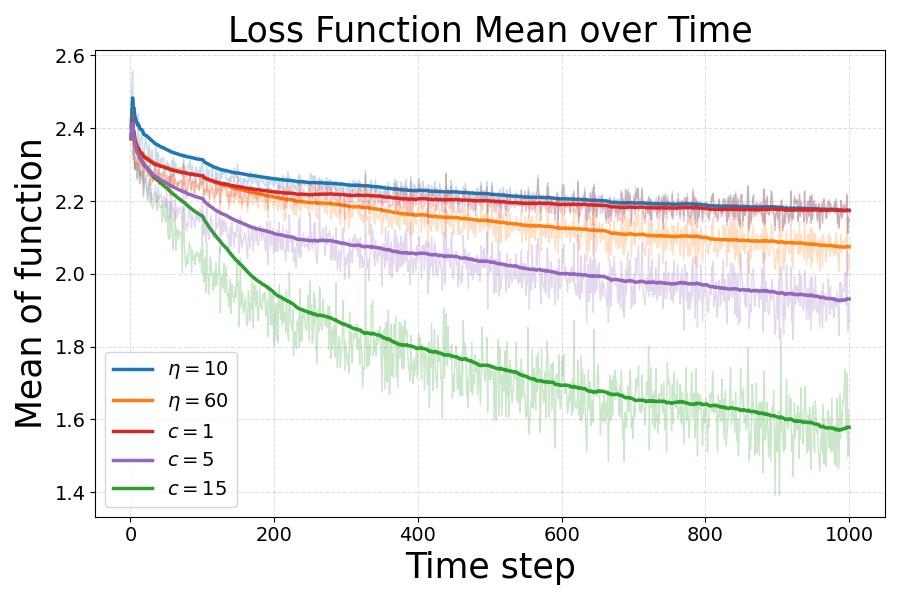}
        \caption{\small{ResNet-18 on CIFAR-10}}
        \label{fig:c_sensitivity_perf_resnet}
    \end{subfigure}
    \caption{\small{Sensitivity of \algoname\ to the hyperparameter $c$. (a) Mean squared gradient in the one-dimensional experiment. Several values of $c$ achieve strong convergence, showing that the method is not sensitive to a single value. (b) Training loss for ResNet-18 on CIFAR-10. Larger values of $c$ improve performance in this high-dimensional setting by keeping the algorithm in the adaptive regime for longer.}}
    \label{fig:c_sensitivity_perf}
\end{figure*}

\begin{figure*}[t]
    \centering
    \begin{subfigure}[t]{0.48\textwidth}
        \centering
        \includegraphics[width=\linewidth]{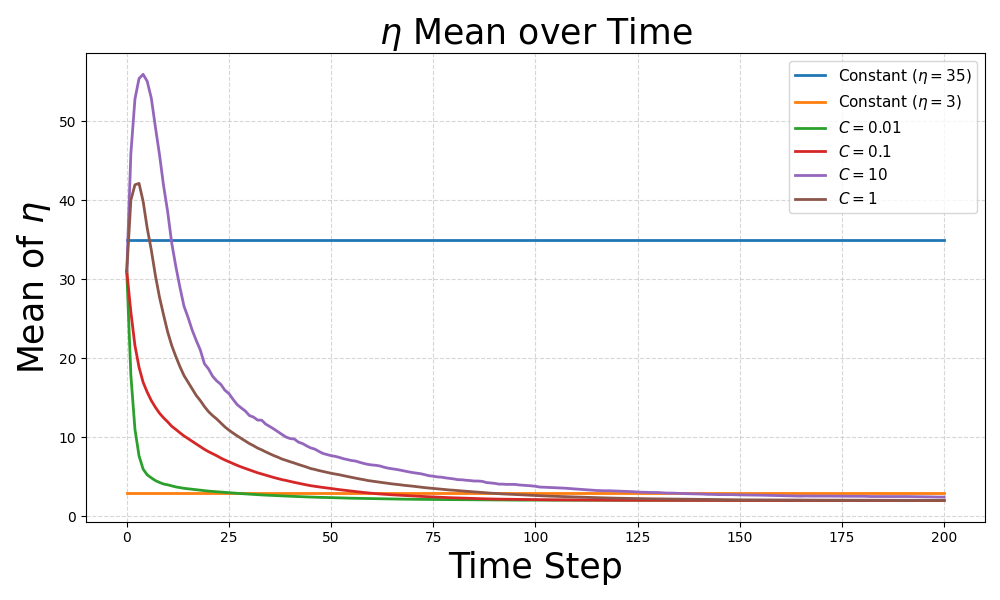}
        \caption{\small{One-dimensional synthetic.}}
        \label{fig:c_sensitivity_eta_1d}
    \end{subfigure}
    \hfill
    \begin{subfigure}[t]{0.48\textwidth}
        \centering
        \includegraphics[width=\linewidth]{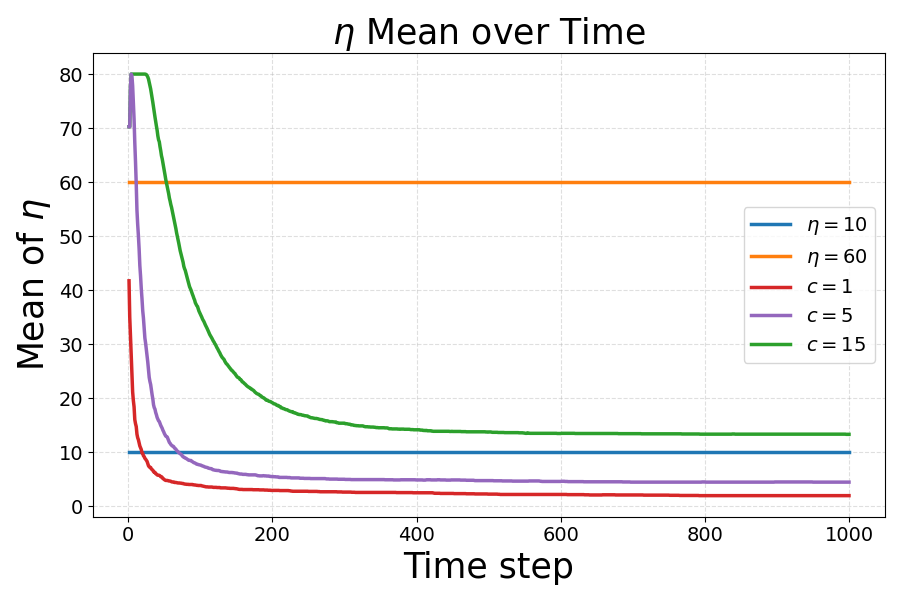}
        \caption{\small{ResNet-18 on CIFAR-10}}
        \label{fig:c_sensitivity_eta_resnet}
    \end{subfigure}
    \caption{\small{Evolution of the acceptance threshold $\eta_t$ for different values of $c$. (a) In the one-dimensional experiment, smaller $c$ values reduce $\eta_t$ more aggressively, while larger values keep the algorithm adaptive for longer. (b) In the ResNet-18 experiment, increasing $c$ leads to larger thresholds during the early phase, which enables more frequent accepted updates before the threshold is gradually tightened.}}
    \label{fig:c_sensitivity_eta}
\end{figure*}

Figures~\ref{fig:c_sensitivity_perf} and~\ref{fig:c_sensitivity_eta} demonstrates the robustness of  \algoname\ to the value of the  hyperparameter $c$ in a wide range.  In the one-dimensional experiment, the choices $c=0.1$, $c=1$, and $c=10$ all achieve strong convergence, although they induce different threshold trajectories. In contrast, $c=0.01$ offers a very slow convergence, even slower than the fixed choice of $\eta=35$. This is due to the fact that, at very early iterations of the algorithm, we have $\sigma_{\min}^2> c \|\widetilde{\mathbf{G}}_t\|_2^2$, and hence, $\eta$ will be set to $\eta_{\min}$ very early,  leading to very infrequent acceptance of computations. In the acceptable range of $c$,  the smaller value $c=0.1$ tightens the acceptance rule more quickly, while the larger value $c=10$ keeps $\eta_t$ large for longer before gradually reducing it. As can be seen from the figure, any choice of $c$ in the range of $0.1$-$10$ leads a fast convergence. In the ResNet-18 experiment on CIFAR-10, increasing $c$ from $1$ to $5$ and $15$ improves performance in this setting, because the larger tolerance allows the algorithm to remain in the adaptive regime for more iterations and make faster progress before tightening the acceptance rule.

These results suggest that $c$ can be selected by a simple calibration procedure rather than by an exhaustive search. One practical approach is to perform a small logarithmic sweep over candidate values and choose a value that produces a stable threshold trajectory: $\eta_t$ should be large enough at the beginning to permit frequent progress, but should decrease over time as the optimization approaches a stationary region. Equivalently, one can monitor $\eta_t$ and the acceptance frequency during a short pilot run. If $\eta_t$ reaches $\eta_{\min}$ too early, then $c$ is likely too small; if $\eta_t$ remains large for too long and the loss or gradient norm plateaus at a high error floor, then $c$ is likely too large. An adaptive choice of $c$ is a natural extension and is left for future work.

Overall, $c$ provides a flexible mechanism for controlling the tradeoff between update frequency and update quality. The ablation results show that \algoname\ remains effective across a meaningful range of values of $c$.

\end{document}